\documentclass{article} 
\usepackage{iclr2022_conference,times}

\usepackage{amsmath,amsfonts,bm}
\usepackage{hyperref}
\usepackage{url}
\usepackage{xcolor}         
\usepackage{graphicx}
\usepackage{wrapfig}
\usepackage{enumitem}
\usepackage[ruled,vlined,linesnumbered]{algorithm2e}
\usepackage{algcompatible}
\usepackage{multirow}
\title{A composable autoencoder-based iterative algorithm for accelerating numerical simulations}

\author{Rishikesh Ranade, Chris Hill, Haiyang He, Amir Maleki, Norman Chang \& Jay Pathak \\
CTO Office\\
Ansys Inc\\
Canonsburg, PA 15317, USA \\ }

\iclrfinalcopy 
\begin{document}

\maketitle

\begin{abstract}
Numerical simulations for engineering applications solve partial differential equations (PDE) to model various physical processes. Traditional PDE solvers are very accurate but computationally costly. On the other hand, Machine Learning (ML) methods offer a significant computational speedup but face challenges with accuracy and generalization to different PDE conditions, such as geometry, boundary conditions, initial conditions and PDE source terms. In this work, we propose a novel ML-based approach, CoAE-MLSim (\textbf{Co}mposable \textbf{A}uto\textbf{E}ncoder \textbf{M}achine \textbf{L}earning \textbf{Sim}ulation), which is an unsupervised, lower-dimensional, local method, that is motivated from key ideas used in commercial PDE solvers. This allows our approach to learn better with relatively fewer samples of PDE solutions. The proposed ML-approach is compared against commercial solvers for better benchmarks as well as latest ML-approaches for solving PDEs. It is tested for a variety of complex engineering cases to demonstrate its computational speed, accuracy, scalability, and generalization across different PDE conditions. The results show that our approach captures physics accurately across all metrics of comparison (including measures such as results on section cuts and lines).
\end{abstract}

\section{Introduction}\label{introduction}

Numerical solutions to partial differential equations (PDEs) are dependent on PDE conditions such as, geometry of the computational domain, boundary conditions, initial conditions and source terms. Commercial PDE solvers have shown a tremendous success in accurately modeling PDEs for a wide range of applications. These solvers generalize across different PDE conditions but can be computationally slow. Moreover, their solutions are not reusable and need to be solved from scratch every time the PDE conditions are changed. 

The idea of using Machine Learning (ML) with PDEs has been explored for several decades \citep{crutchfield1987equations, kevrekidis2003equation} but with recent developments in computing hardware and ML techniques, these efforts have grown immensely. Although ML approaches are computationally fast, they fall short of traditional PDE solvers with respect to accuracy and generalization to a wide range of PDE conditions. Most data-driven and physics-constrained approaches employ static-inferencing strategies, where a mapping function is learnt between PDE solutions and corresponding conditions. In many cases, PDE conditions are sparse and high-dimensional, and hence, difficult to generalize. Additionally, current ML approaches do not make use of the key ideas from traditional PDE solvers such as, domain decomposition, solver methods, numerical discretization, constraint equations, symmetry evaluations and tighter non statistical evaluation metrics, which were established over several decades of research and development. In this work, we propose a novel ML approach that is motivated from such ideas and relies on dynamic inferencing strategies that afford the possibility of seamless coupling with traditional PDE solvers, when necessary.

The proposed ML-approach, CoAE-MLSim, which is a \textbf{Co}mposable \textbf{A}uto\textbf{E}ncoder \textbf{M}achine \textbf{L}earning \textbf{Sim}ulation algorithm, operates at the level of local subdomains, which consist of a group of pixels in $2$D or voxels in $3$D (for example $8$ or $16$ in each spatial direction).
\begin{wrapfigure}{r}{0.5\textwidth}
    \centering
    \vspace{-1em}
    \includegraphics[width=6cm]{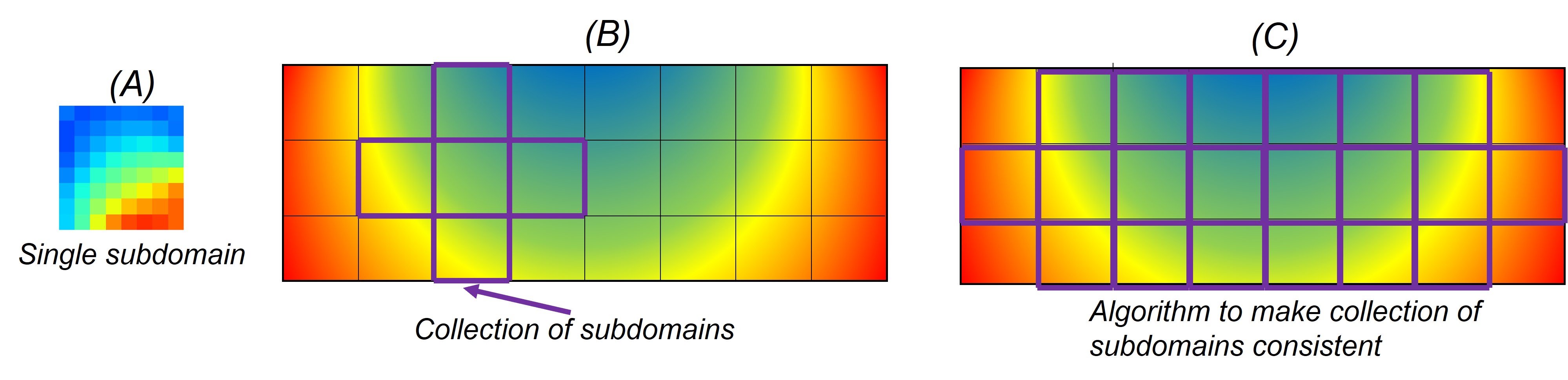}
    \vspace{-3ex}
    \caption{CoAE-MLSim components}
    \label{fig:1}
\end{wrapfigure}
CoAE-MLSim has $3$ main components: A) learn solutions on local subdomains, B) learn the rules of how a group of local subdomains connect together to yield locally consistent solutions \& C) deploy an iterative algorithm to establish local consistency across all groups of subdomains in the entire computational domain. The solutions on subdomains are learnt into low-dimensional representations using solution autoencoders, while the rules between groups of subdomains, corresponding to flux conservation between PDE solution is learnt using flux conservation autoencoders. The autoencoders are trained \textit{a priori} on local subdomains using very few PDE solution samples. During inference, the pretrained autoencoders are combined with iterative solution algorithms to determine PDE solutions for different PDEs over a wide range of sparse, high-dimensional PDE conditions. The solution strategy in the CoAE-MLSim approach is very similar to traditional PDE solvers, moreover the iterative inferencing strategy allows for coupling with traditional PDE solvers to accelerate convergence and improve accuracy and generalizability.

\textbf{Significant contributions of this work:} 
\begin{enumerate}
    \item The CoAE-MLSim approach combines traditional PDE solver strategies with ML techniques to accurately model numerical simulations.
    \item Our approach operates on local subdomains and solves PDEs in a low-dimensional space. This enables scaling to arbitrary PDE conditions and obtain better generalization. 
    \item The CoAE-MLSim approach training corresponds to training the various autoencoders on local subdomains and hence, it contains very few parameters and requires less data. 
    \item The iterative inferencing algorithm is completely unsupervised and allows for coupling with traditional PDE solvers.
    \item Finally, we show generalization of our model for wide variations in sparse, high dimensional PDE conditions using evaluation metrics in tune with commercial solvers.
\end{enumerate}

\section{Related works}\label{related works}

\textbf{PDE Operator learning}: Recently, there has been a lot of focus on learning mesh independent, infinite dimensional operators with neural networks (NNs) \cite{bhattacharya2020model, anandkumar2020neural, li2020multipole, patel2021physics, lu2021learning, li2020fourier}. The neural operators are trained from high-fidelity solutions generated by traditional PDE solvers on a mesh of specific resolution and do not require any knowledge of the PDE. For dynamical systems, the neural operators have shown reasonable extrapolation capabilities, however, the generalizability of these methods to different PDE conditions remains to be seen. 

\textbf{Physics constrained optimization}: Research in this space involves constraining neural networks with additional physics-based losses introduced through loss re-engineering. PDE constrained optimization have shown to improve interpretability and accuracy of results. \citet{raissi2019physics} and \citet{raissi2018hidden} introduced the framework of physics-informed neural network (PINN) to constrain neural networks with PDE derivatives computed using Automatic Differentiation (AD) \cite{baydin2018automatic}. In the past couple of years, the PINN framework has been extended to solve complicated PDEs representing complex physics \citep{jin2021nsfnets, mao2020physics, rao2020physics, wu2018physics, qian2020lift, dwivedi2021distributed, haghighat2021physics, haghighat2021sciann, nabian2021efficient, kharazmi2021hp, cai2021flow, cai2021physics, bode2021using, taghizadeh2021explicit, lu2021deepxde, shukla2021parallel, hennigh2020nvidia, li2021kohn}. More recently, alternate approaches that use discretization techniques using higher order derivatives and specialize numerical schemes to compute derivatives have shown to provide better regularization for faster convergence \citep{ranade2021discretizationnet, gao2021phygeonet, wandel2020learning, he2020unsupervised}.

\textbf{Mesh-based learning}: \citet{battaglia2018relational} introduced a graph network architecture, which has proven to be effective in solving dynamical systems directly on unstructured computational meshes. \citet{sanchez2020learning} and \citet{pfaff2020learning} use the graph network for robustly solving transient dynamics on arbitrary meshes and accurately capture the transient solution trajectories.

\textbf{Differentiable solver frameworks for learning PDEs}: Training NNs within differentiable solver frameworks has shown to improve learning and provide better control of PDE solutions and transient system dynamics \citep{amos2017optnet, um2020solver, de2018end, toussaint2018differentiable, wang2020differentiable, holl2020learning, portwood2019turbulence}. Differentiable simulations are useful in providing rapid feedback to neural networks to improve convergence stability and enable efficient exploration of solution space. \citet{bar2019learning}, \citet{xue2020amortized} and \citet{kochkov2021machine} train neural networks in conjunction with a differentiable solver to learn the PDE discretization on coarse grids. \citet{singh2017augmentation} and \citet{ holland2019field} use differentiable solvers to tune model parameters in a simulation.

\textbf{Local and latent space learning}: Local learning from smaller restricted domains have proved to accelerate learning of neural networks and provide better accuracy and generalization. \citet{lu2021one} and \citet{wang2021train} learn on localized domains but infer on larger computational domains using a stitching process. \citet{bar2019learning} and \citet{kochkov2021machine} learn coefficients of numerical discretization schemes from high fidelity data, which is sub-sampled on coarse grids. \citet{beatson2020learning} learns surrogate models for smaller components to allow for cheaper simulations. Other methods compress PDE solutions on to lower-dimensional manifolds. This has shown to improve accuracy and generalization capability of neural networks \citep{wiewel2020latent, maulik2020reduced, kim2019deep, murata2020nonlinear, fukami2020convolutional, ranade2021latent}.

\section{CoAE-MLSim model details}\label{pdeml approach}

\subsection{Similarities with traditional PDE solvers}
Consider a set of coupled PDEs with $n$ solution variables. For the sake of notation simplicity, we take $n=2$, such that $u(x,y,z,t)$ and $v(x,y,z,t)$ are defined on a computational domain $\Omega$ with boundary conditions specified on the boundary of the computational domain, $\Omega_b$. It should be noted that extension to more solution variables is trivial.
The coupled PDEs are defined as follows:
\begin{equation}
        L_1(u, v) - F_1  = 0; L_2(u, v) - F_2 = 0
    \label{eq1}
\end{equation}
where, $L_1,$ $L_2$ denote PDE operators and $F_1,$ $F_2$ represent PDE source terms. The PDE operators can vary for different PDEs. For example, in a non-linear PDE such as the unsteady, incompressible Navier-Stokes equation the operator, $L = \frac{\partial}{\partial t} + \vec{a}.\vec{\nabla} - \vec{\nabla}.\vec{\nabla}$

Traditional PDE solvers solve PDEs given in Eq. \ref{eq1} by representing solutions variables, $u, v$, and their linear and non-linear spatio-temporal derivatives on a discretized computational domain. The numerical approximations on discrete domains are computed on a finite number of computational elements known as a mesh, using techniques such as Finite Difference Method (FDM), Finite Volume Method (FVM) and Finite Element Method (FEM). These solvers use iterative solution algorithms to conserve fluxes between neighboring computational elements and determine consistent PDE solutions over the entire domain at convergence. The CoAE-MLSim approach is designed to perform similar operations but at the level of subdomains with assistance from ML techniques.

\subsection{CoAE-MLSim for steady-state PDEs} \label{steady PDEs}
\begin{figure}[h]
  \centering
  \includegraphics[width=\textwidth]{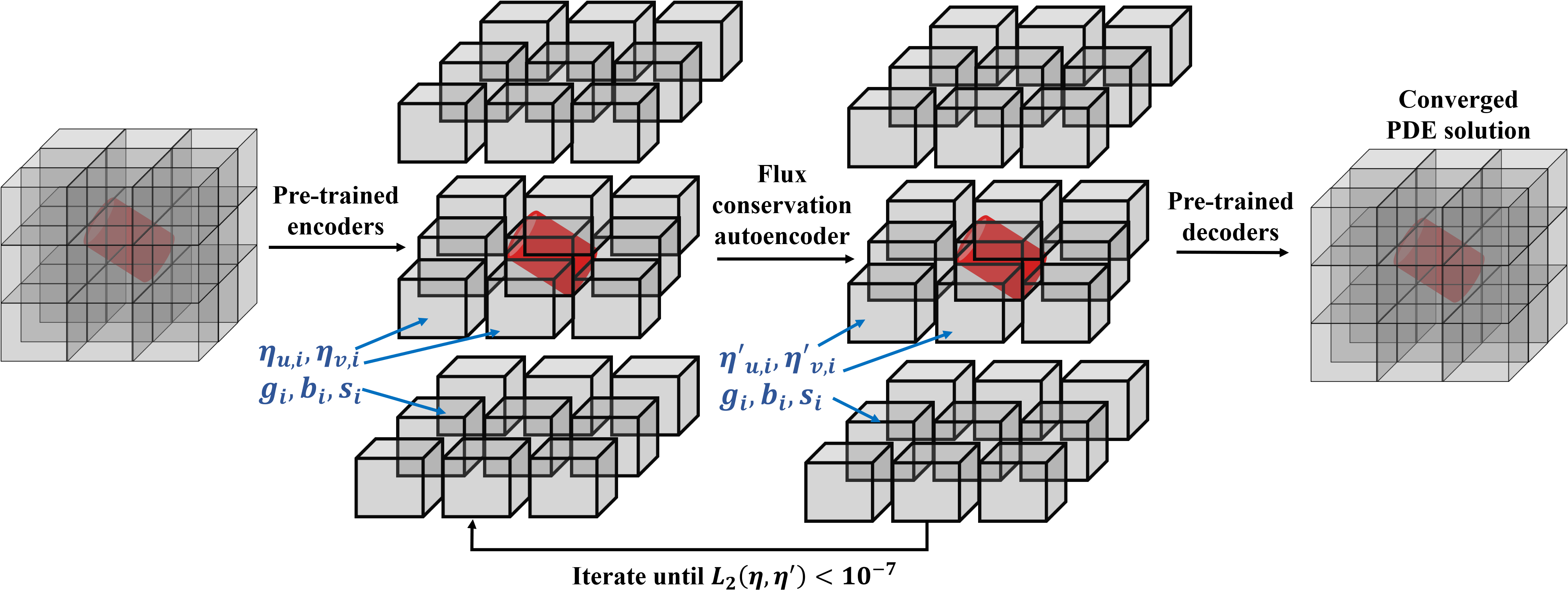}
  \caption{Steady state CoAE-MLSim algorithm}
  \label{steady-state}
\end{figure}
Steady-state PDEs correspond to partial differential equations without time dependencies. The PDE solutions in these equations are primarily governed by PDE conditions, which can have high-dimensional and sparse representations. The solution algorithm of CoAE-MLSim approach for steady-state PDEs is shown in Fig. \ref{steady-state} and Alg. \ref{alg:predict}. Similar to traditional solvers, the CoAE-MLSim approach discretizes the computational domain into several subdomains. Each subdomain has a constant physical size and represents both PDE solutions and conditions. For example, a subdomain cutting across the cylinder in Fig. \ref{steady-state} represents the geometry of the part-cylinder. In a steady-state problem, the computational domain is initialized with initial PDE solutions that are either randomly sampled or generated by coarse-grid PDE solvers. Pre-trained encoders are used to encode the initial solutions as well as user-specified PDE conditions into lower-dimensional latent vectors, $\eta_u, \eta_v$ corresponding to PDE solutions and $g, b, s$ corresponding to geometry, boundary conditions and sourceterms, respectively. The flux conservation is iteratively established by passing batches of neighboring subdomain latent vectors through the flux conservation autoencoder. In each iteration, the algorithm loops over all the subdomains, gathers neighbors for a given subdomain and concatenates the neighboring latent vectors. The concatenated latent vectors are evaluated through the flux conservation autoencoder to obtain the new solution latent vector ($\eta$'$_{\vec{u}}$, $\eta$'$_{\vec{v}}$) state on each subdomain. Subdomains on the boundary have fewer neighbors and latent vectors are zero padded in such cases. The flux conservation autoencoder couples all the solution variables with PDE conditions to ensure that all the dependencies are captured. The iteration stops when the $L_2$ norm of change in solution latent vectors meets a specified tolerance, otherwise the latent vectors are updated and the iteration continues. The encodings of the PDE conditions are not updated and help in steering the solution latent vectors to an equilibrium state that is decoded to PDE solutions using pre-trained decoders $(g)$ on the computational domain. The iterative procedure used in the CoAE-MLSim approach can be implemented using several linear or non-linear equation solvers, such as Fixed point iterations \citep{bai2019deep}, Gauss Siedel, Newton's method etc., that are used in commercial PDE solvers. Physics constrained optimization at inference time can be used to improve convergence robustness and fidelity with physics.
\begin{algorithm}[H]
\caption{Steady state solution methodology of CoAE-MLSim approach}
\label{alg:predict}
Domain Decomposition: Computational domain $\Omega$ $\rightarrow$ Subdomains $\Omega_s$

Initialize solution on all $\Omega_s$: $\vec{u}(x) = 0.0$ for all $x \in \Omega_s$

Encode PDE solution and conditions on all $\Omega_s$: $\eta_{\vec{u}} = e_u(\vec{u(\Omega_s)}), \eta_g = e_g(g(\Omega_s)), \eta_b = e_b(b(\Omega_s)), \eta_s = e_s(s(\Omega_s))$

$\epsilon_t = 1e^{-4}$ \\
\While{$\epsilon$ > $\epsilon_t$}{
  \For{$\Omega_s \in \Omega$}{
        \text{Gather neighbors of}  $\Omega_s$: $\Omega_{nb} = [\Omega_s, \Omega_{left}, \Omega_{right}, ...]$\\
        $\eta_{\vec{u}}' = \Theta(\eta^{nb}_{\vec{u}}, \eta^{nb}_b, \eta^{nb}_g, \eta^{nb}_s)$
}
Compute $L_2$ norm: $\epsilon = ||\eta_{\vec{u}}-\eta_{\vec{u}}'||_2$\\
Update: $\eta_{\vec{u}} \leftarrow \eta_{\vec{u}}'$ \text{for all} $\Omega_s \in \Omega$
}
Decode PDE solution on all $\Omega_s$: $\vec{u} = g_u(\eta_{\vec{u(\Omega_s)}})$

\end{algorithm}

\subsection{CoAE-MLSim for transient PDEs} \label{transient PDEs}
\begin{figure}[h]
  \centering
  \includegraphics[width=\textwidth]{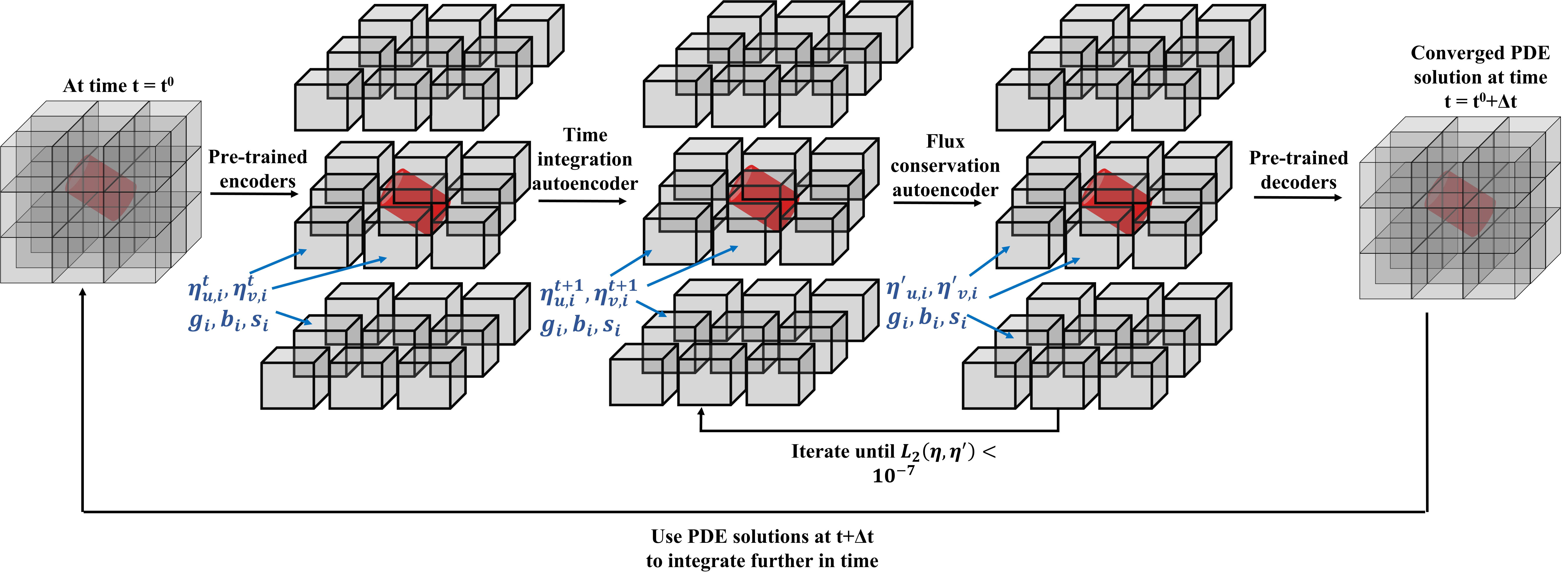}
  \caption{Transient CoAE-MLSim algorithm}
  \label{transient-state}
\end{figure}
The algorithm for transient CoAE-MLSim differs slightly from steady-state. Our approach models transient PDEs using two methods, tightly coupled and loosely coupled. The tightly coupled is analogous to the steady-state algorithm where time is considered as an additional dimension in addition to space and modeled using the algorithm described in Alg. \ref{alg:predict}. On the other hand, in the loosely coupled approach the spatial and transient effects are decoupled. The decoupling allows for better modeling of long range time dynamics and results in improved stability and generalizability. The solution methodology shown in Figure \ref{transient-state} corresponds to the loosely coupled approach and differs from the steady-state methodology in Figure \ref{steady-state} in that it uses an additional time integration autoencoder, which integrates the solution latent vectors in time. More details on the time integration autoencoder are provided in Section \ref{time integration}. Analogous to traditional PDE solvers, a flux conservation is applied after every time integration step. This is very important in establishing local consistency between neighborhood subdomains and minimizing error accumulation resulting from the transient process.
\subsection{PDE solution and condition autoencoders} \label{autoencoders}
Autoencoders are used to establish lower-dimensional latent vectors, $\eta, \eta_s, \eta_g$, for both PDE solutions and conditions represented on local subdomains and reconstruct them accurately for a given PDE.
\begin{wrapfigure}{r}{0.65\textwidth}
    \centering
    \vspace{-1em}
    \includegraphics[width=0.65\textwidth]{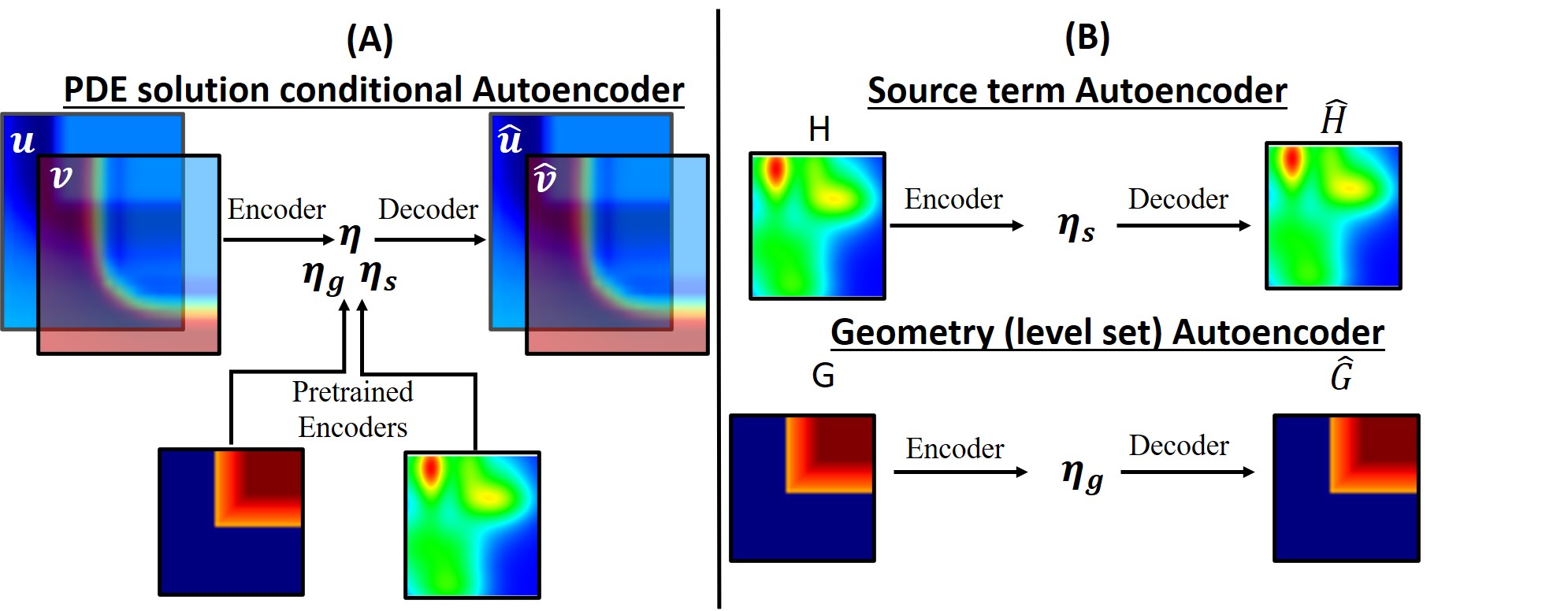}
    \vspace{-3ex}
    \caption{Autoencoders PDE solutions and condition}
    \label{solution autoencoder}
\end{wrapfigure}
Figure \ref{solution autoencoder} shows a schematic of the autoencoder setup used in the CoAE-MLSim \citep{ranade2021latent}. Although the representative subdomains shown are $2$-D, same concepts apply in higher dimensions. The geometry autoencoders encode a representation of the geometry into latent vector, $\eta_g$. In this work, we adopt a Signed Distance Field (SDF) representation of the geometry because it is smooth and differentiable \citep{maleki2021geometry}. Similarly, the PDE source terms are encoded to their respective latent vector, $\eta_s$. The solution autoencoders are conditioned upon the compressed latent vectors of the PDE conditions by concatenating them with the latent vector ($\eta$) of the PDE solutions. Each solution variable can be trained using a different autoencoder to improve accuracy. The autoencoders are trained using true PDE solutions generated for random PDE conditions. These solutions are generated on entire computational domains and then divided into smaller subdomains for training.

\subsection{Flux conservation autoencoder} \label{flux conservation}
\begin{wrapfigure}{r}{0.65\textwidth}
    \centering
    \vspace{-1em}
    \includegraphics[width=0.65\textwidth]{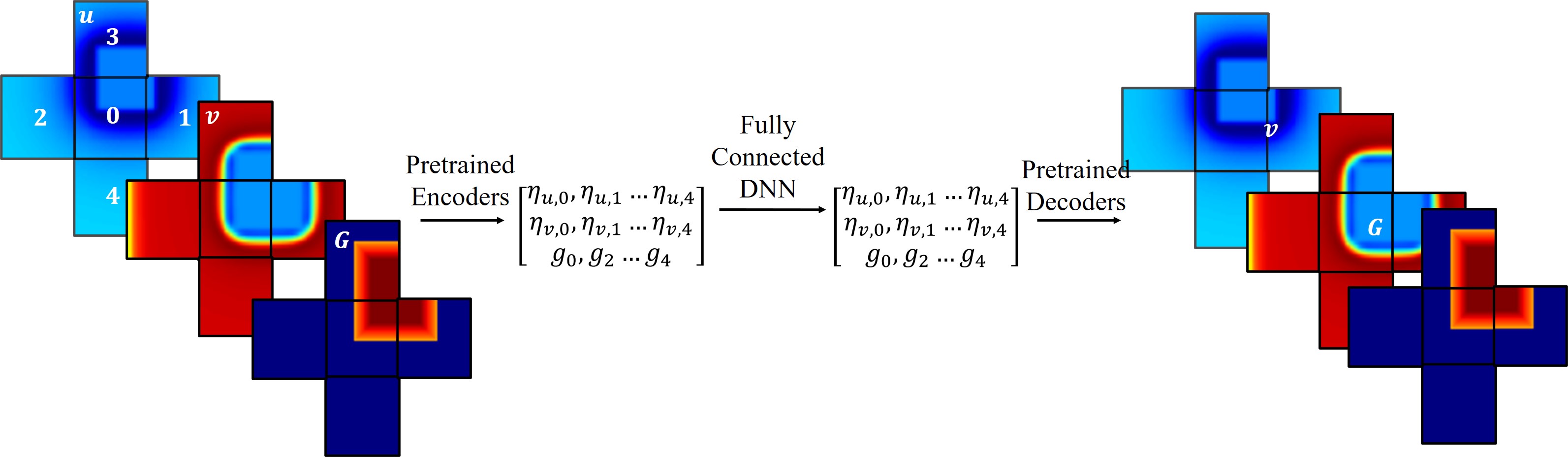}
    \vspace{-3ex}
    \caption{Flux conservation autoencoder}
    \label{flux conservation autoencoder}
\end{wrapfigure}
Flux conservation autoencoders are the primary work horse of the solution algorithm and are responsible for a bulk of the calculations. Their main function is to disperse solution information throughout the computational domain by transferring information between neighborhood subdomains and from boundaries and geometries. Figure \ref{flux conservation autoencoder} shows the schematic of the flux conservation autoencoder, which operates on the latent vectors of PDE solutions and conditions on a group of neighboring subdomains. The inputs and outputs to this network consist of concatenated latent vectors of all solution variables and PDE conditions on a group of neighboring subdomains. 
It uses a deep fully connected neural network to learn these relationships. The samples generated for autoencoders in Section \ref{autoencoders} are used for training the flux conservation autoencoders as well.

\subsection{Time integration autoencoder} \label{time integration}

\begin{wrapfigure}{r}{0.65\textwidth}
    \centering
    \vspace{-2em}
    \includegraphics[width=0.65\textwidth]{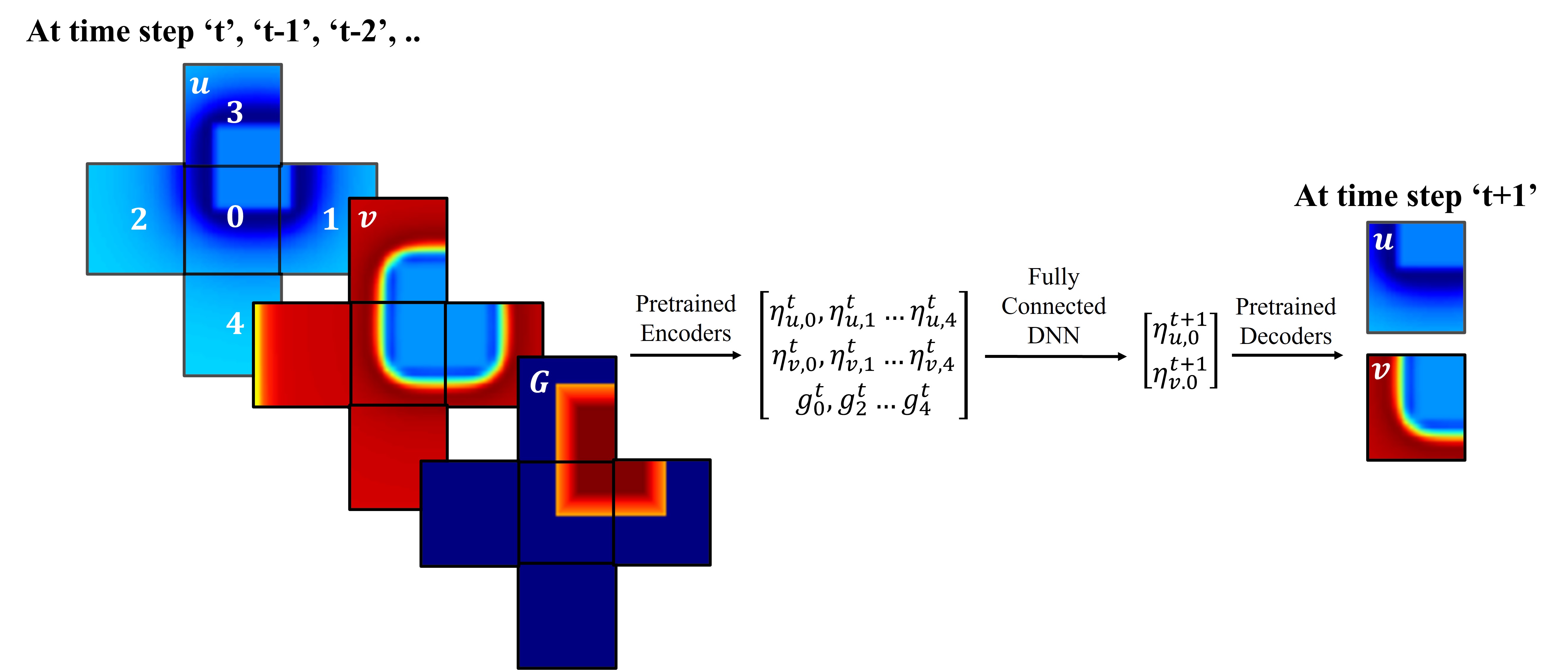}
    \vspace{-3ex}
    \caption{Time integration autoencoder}
    \label{time integration autoencoder}
\end{wrapfigure}

Figure \ref{time integration autoencoder} shows the schematic of the time integration autoencoder, which operates on the latent vectors of PDE solutions and conditions on a group of neighboring subdomains. 
The time integration autoencoder uses fully connected networks to transform the solution latent vector of the center subdomain (corresponding to $0$ in Fig. \ref{time integration autoencoder}) in time. 

The input of the time integration autoencoder can stack latent vectors from multiple previous time steps, $t, t-1, t-2 ...$ , to predict the latent vectors for the next time step, $t+1$. Each solution variable is trained using a different autoencoder for improved accuracy. Similar to other autoencoders, the time integration autoencoder also learns from randomly generate solution samples. Details regarding network architectures and training mechanics of all the autoencoder networks are provided in the appendix.

\subsection{Why Autoencoders?}
Solutions to classical PDEs such as the Laplace equation can be represented by homogeneous solutions as follows:
\begin{equation}
        \phi(x,y) = a_0 + a_1x + a_2y + a_3(x^2-y^2) + a_4(2xy) + ...
    \label{eq2}
\end{equation}
where, $\vec{A} = {a_0, a_1, a_2, ..., a_n}$ are constant coefficients that can be used to reconstruct the PDE solution on any local subdomain. $\vec{A}$ can be considered as a compressed encoding of the Laplace solutions. Since, it is not possible to explicitly derive such compressed encodings for other high dimensional and non-linear PDEs, the CoAE-MLSim approach relies on autoencoders to extract latent vectors for such PDEs from solution data. It is known that non-linear autoencoders with good compression ratios can learn powerful non-linear generalizations \citep{goodfellow2016deep, rumelhart1985learning, bank2020autoencoders}. Thus, autoencoders enable efficient latent space computations. Additionally, autoencoders have great denoising abilities, which improve robustness and stability, when used in iterative settings \citep{ranade2021latent}. Finally, autoencoders are data-efficient and result in a small number of learnable model parameters and much faster training. For these reasons, autoencoders form the fundamental building block of the CoAE-MLSim approach.  

\subsection{Attributes of CoAE-MLSim solution methodology}

\begin{enumerate}
    \item \textbf{Unsupervised algorithm:} It is important to note that the CoAE-MLSim solution approach is unsupervised. Although, the autoencoders are trained on true PDE solutions generated for random PDE conditions, the iterative solution procedure described in Section \ref{steady PDEs} is never explicitly taught the process of computing PDE solutions and discovers solutions with a minimal knowledge about the rules of local consistency. This is remarkably similar to how traditional PDE solvers would operate.
    \item \textbf{Local learning:} Since the autoencoders are trained on local subdomains, they have fewer trainable parameters and need very less data samples to learn the dynamics between PDE solutions and conditions. In all the use cases presented in this work, $100$-$1000$ PDE solutions are generated for random PDE conditions, which span a high dimensional space.
    \item \textbf{Latent space representation:} The iterative solution procedure is carried out in a compressed latent space to achieve solution speed-ups. Furthermore, compressed representations of sparse, high-dimensional PDE conditions improves generalizability. 
    \item \textbf{Coupling with traditional PDE solvers:} The iterative inferencing strategy allows for coupling with traditional PDE solvers. In this work, we have explored the possibility of initializing the steady-state CoAE-MLSim approach with coarse mesh PDE solutions generated by true PDE solvers to achieve improved speed up and accuracy on fine resolution meshes.
\end{enumerate}

\section{Experiments and Results}\label{results}
In this section, we demonstrate the CoAE-MLSim approach for five use cases, steady-state Laplace equation, steady-state conjugate heat transfer, transient vortex decay and transient flow over a cylinder. Additional details corresponding to these use cases are presented in the appendix sections. All the experiments have varying levels of complexity across geometries, boundary and initial conditions \& source terms imposed on the PDE. The source code and the datasets used in our experiments and analysis can be made available upon acceptance.

\textbf{Data generation and training mechanics}
The data required to train the several autoencoders in the CoAE-MLSim approach is generated using Ansys Fluent \citep{fluent2015ansys}, except for a few cases where publicly available data sets are used. As stated earlier, the data requirements are minimal and each case requires about $100$-$1000$ solutions, depending on the complexity of physics and dimensionality of PDE conditions. The training in the CoAE-MLSim corresponds to training several autoencoders. The network architectures and training mechanics are general and described in appendix \ref{appendix:training}. 

\subsubsection{Steady State: Laplace equations} \label{appendix:laplace}

The Laplace equation is defined as follows:
\begin{equation} \label{laplace_eq}
    \nabla^2 \phi (\vec{x}) = 0.0 
\end{equation}
subjected to a Dirichlet boundary condition, $\phi (\vec{x_b}) = f_b$ or a Neumann boundary condition, $\frac{\partial \phi}{\partial \vec{x_b}} = f_b$. 

\begin{figure}[h]
  \centering
  \includegraphics[width=0.55\textwidth]{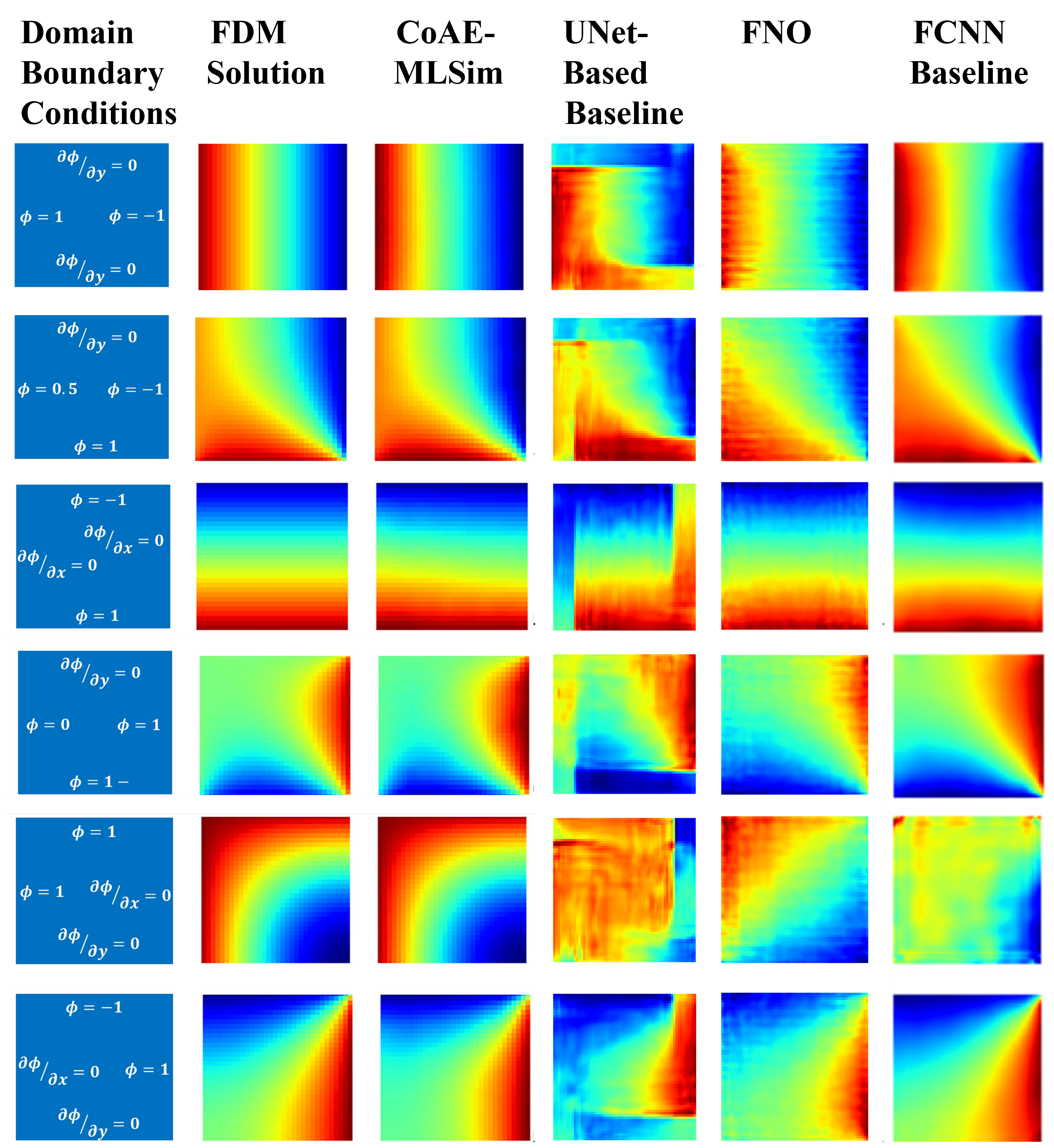}
  \caption{Laplace equation contour results}
  \label{laplace_res}
\end{figure}

The Laplace equation is a second order, linear PDE that represents a canonical problem for benchmarking linear solvers. It is solved on a square computational domain with a resolution of $64$x$64$ computational elements for random boundary conditions sampled from Dirichlet and Neumann. Here, we provide visual comparisons between the CoAE-MLSim approach and a second-order finite difference method (FDM) approach in Fig. \ref{laplace_res}. The PDE solutions are locally scaled with the Dirichlet boundary condition. We compare the CoAE-MLSim with $3$ different ML approaches, Unet \citep{ronneberger2015u}, FNO \citep{li2020fourier} and FCNN. The training data for CoAE-MLSim corresponds to $300$ solutions generated for arbitrary boundary conditions using a second order finite difference method, whereas all the baselines are trained with $6400$ solutions. Details of the baseline models as well as other information regarding our approach is presented in the appendix section, \ref{appendix:laplace}.

It may be observed from Fig. \ref{laplace_res} that the CoAE-MLSim approach outperforms the baseline ML models. The mean absolute errors for the CoAE-MLSim approach over $50$ unseen testing samples is $7e^{-3}$. On the other hand, Unet, FNO and FCN have mean absolute errors of $1.65e^{-1}$, $1.9e^{-1}$ and $3e^{-2}$ respectively. All errors are with respect to the second order FDM solution. It may be observed that the FCNN performs better than both UNet and FNO and this points to an important aspect about representation of PDE conditions and its impact on accuracy. The representation of boundary conditions on a 64x64 grid is very sparse and high-dimensional, making it very challenging for the networks to learn. On the other hand, the FCNN uses a low-dimensional, dense encoding as an input and hence is able to learn more effectively. Nonetheless, the CoAE-MLSim approach provides the best performance.

\subsection{Steady-state: Industrial use case of electronic-chip thermal cooling}\label{cht}

This experiments demonstrates the steady-state CoAE-MLSim approach. It consists of an electronic chip package surrounded by air and subjected to heat sources with random spatial distributions due to uncertainty in electrical heating. The temperature distribution on the chip is governed by a natural convection process, which is a balance between heating due to heat source and cooling because of flow. The heat sources are sampled from a Gaussian mixture model (up to 25 Gaussians with random mean and variances) and represent a high dimensional space ($4096$) with large variations, thereby making it incredibly hard to generalize across. The training data for this case corresponds to only $300$ solutions generated for random heat sources using Ansys Fluent. 

\begin{figure}[h]
  \centering
  \includegraphics[width=\textwidth]{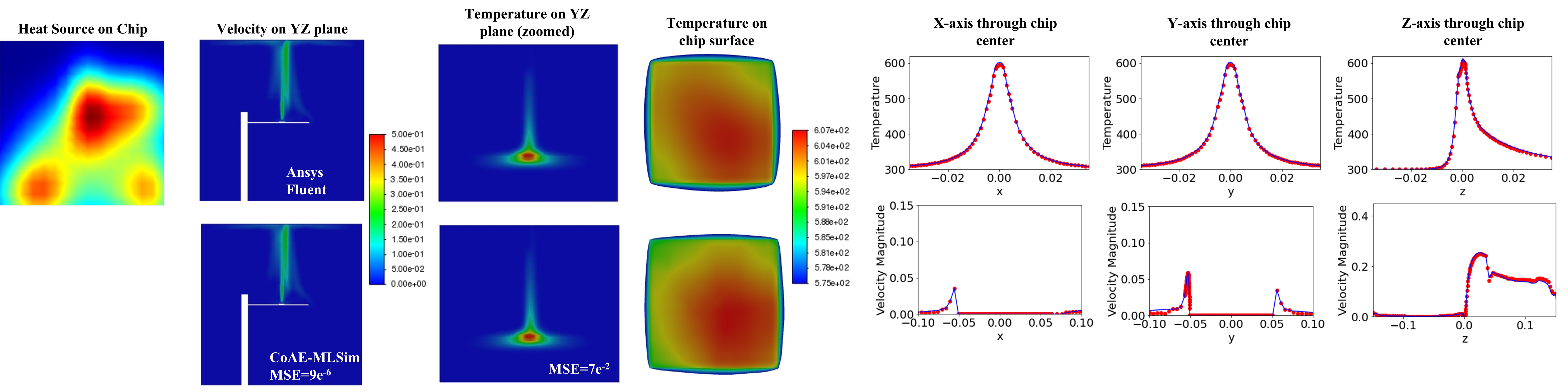}
  \caption{Comparisons: Contour (left) \& Line (right, red: CoAE-MLSim \& blue: Fluent)}
  \label{cht_fig1}
\end{figure}
In Fig. \ref{cht_fig1}, we show the temperature and velocity magnitude contour and line plot comparisons between Ansys Fluent and CoAE-MLSim approach for an unseen heat source. The mean squared error for temperature and velocity magnitude over $50$ unseen test samples is $5.3e^{-2}$ and $6.2e^{-6}$ as compared to Fluent. More details related to the case setup, governing equations, training mechanics, CoAE-MLSim configuration and additional results and analysis is provided in the appendix section \ref{appendix:cht}.

\subsubsection{Steady State: Flow over arbitrary objects} \label{geo}

This experiments demonstrates the steady-state CoAE-MLSim approach. In this experiment, the CoAE-MLSim approach is demonstrated for generalizing across a wide range of geometries. The computational domain consists of a 3-D channel flow over arbitrarily shaped objects. The geometry of these objects is represented with a signed distance field representation and is extremely high dimensional. The Reynolds number equals $20$ and the flow falls within the laminar regime. The objective is to be able to capture the dynamics of flow due to changes in geometry. The training data for this case corresponds to only $300-400$ solutions generated for random geometries using Ansys Fluent. 

\begin{figure}[h]
  \centering
  \includegraphics[width=\textwidth]{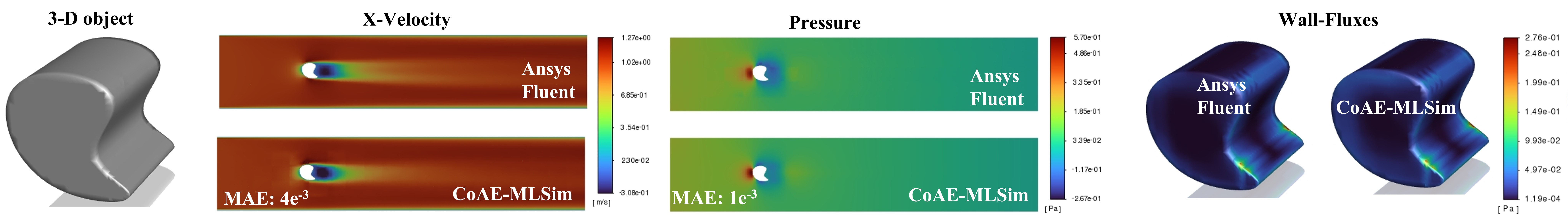}
  \caption{Flow over external object contour plot comparisons}
  \label{arbit_objects}
\end{figure}

Figure \ref{arbit_objects} shows the velocity, pressure and wall flux contour plots comparisons between CoAE-MLSim approach and Ansys Fluent for flow around an unseen arbitrary object. The results match to an acceptable accuracy. In fact, the mean absolute error for pressure and velocity magnitude over $50$ unseen test samples is $2.3e^{-2}$ and $9.4e^{-4}$. More details about the case setup, governing equations, additional results and analysis is provided in the appendix section \ref{appendix:geo}.

\subsection{Transient: Vortex decay over time} \label{vortex}
\begin{figure}[h]
  \centering
  \includegraphics[width=\textwidth]{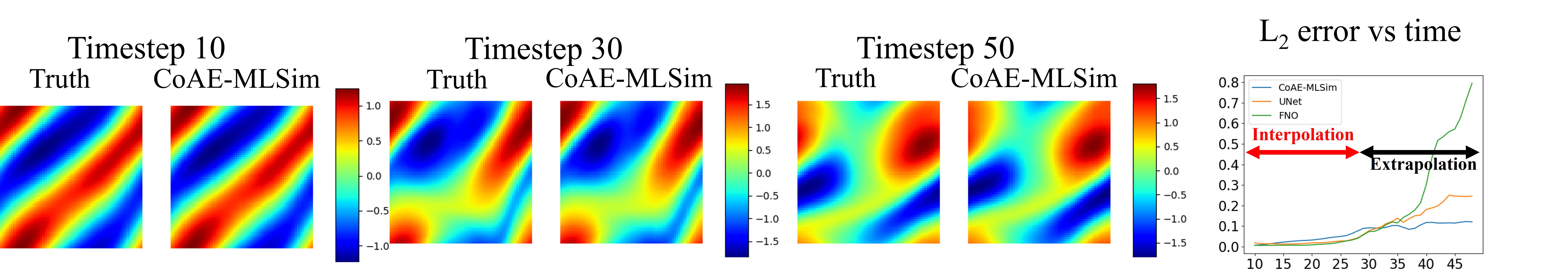}
  \caption{Vortex decay over time with CoAE-MLSim approach}
  \label{vd_1}
\end{figure}
This experiment demonstrates the transient CoAE-MLSim approach. We solve the vorticity form of the Navier-Stokes equation on a 2-D domain. In this case, the transient dynamics of flow are modeled for different choices of initial vorticity. The training data (generate by \cite{li2020fourier}) used to train the CoAE-MLSim autoencoders comprises of $500$ initial conditions and first $25$ timesteps. The testing is carried out on $50$ unseen initial conditions and extrapolated up to $50$ timesteps. It may be observed in Fig. \ref{vd_1} that the CoAE-MLSim predictions match well with the ground truth and the error accumulation is acceptable, especially in the extrapolation region for an unseen test sample. The error accumulation is significantly smaller than ML baselines such as UNet \citep{ronneberger2015u} and Fourier neural operator (FNO) \citep{li2020fourier}. The CoAE-MLSim approach minimizes error accumulation using the flux conservation autoencoder, which enforces local consistency and controls the trajectory of the solution after every time integration. More details about the case setup, governing equations, description of baseline ML models, additional results and analysis is provided in the appendix section \ref{appendix:decay}.

\subsection{Transient: Flow over a cylinder} \label{transient_cy}
\begin{figure}[h]
  \centering
  \includegraphics[width=0.7\textwidth]{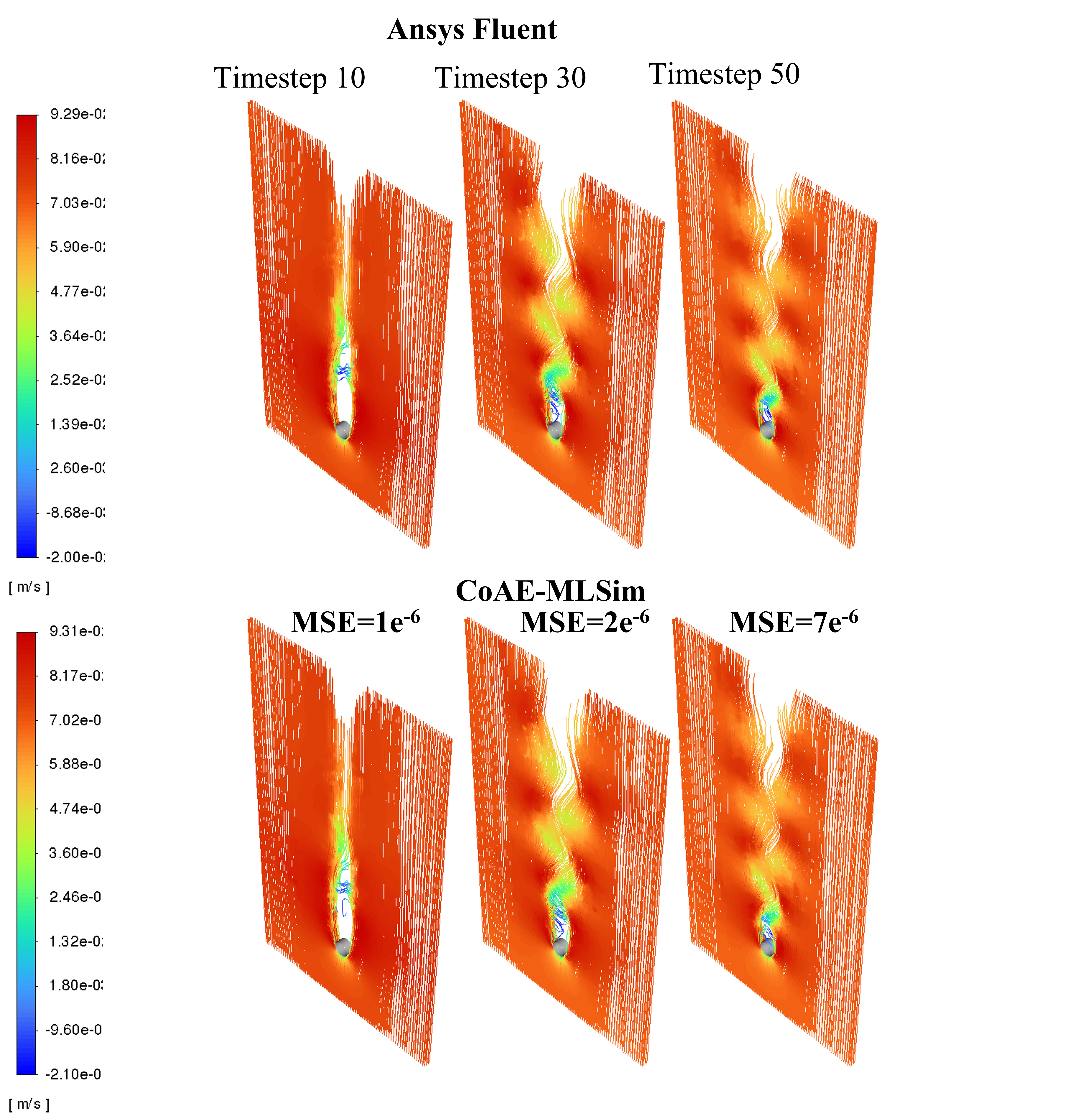}
  \caption{Transient flow comparisons of velocity magnitude}
  \label{cyl_trans_vel}
\end{figure}

Finally, we present a demonstration of the CoAE-MLSim approach in solving the flow around a cylinder problem in a transient setting at a flow Reynolds number equal to $200$, in order to induce unsteady phenomenon in the flow, commonly known as vortex street. In this case, the CoAE-MLSim approach is trained on Reynolds number of $50$ and $1000$ and tested on $200$. The training data corresponds to only $50$ timesteps but the testing is carried out until $100$ timesteps. The timestep used for training is $20$x larger than the one used by Ansys Fluent to generated the training and testing data. The complexity of the problem is increased by adding a constant heat flux to the cylinder, resulting in dissipation of temperature with the flow. Fig. \ref{cyl_trans_vel} shows a comparison of the vector plots of velocity magnitude between Ansys Fluent and our approach. It may be observed that the errors are very small for entire span of time. Additional results are provided in the appendix section \ref{appendix:cylinder_transient}.

\subsection{Discussion}\label{discussion}

\textbf{Ablation study for different subdomain sizes:}

\begin{wrapfigure}{r}{0.3\textwidth}
    \centering
    \vspace{-3em}
    \includegraphics[width=0.25\textwidth]{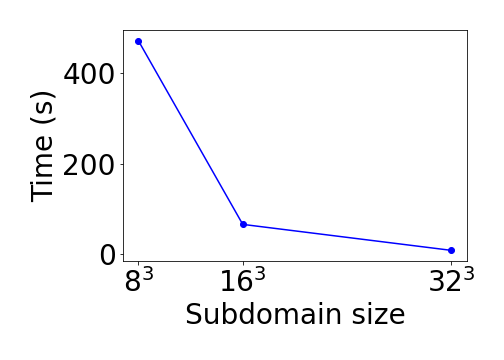}
    \vspace{-2ex}
    \caption{Computation time for different sub-domain sizes}
    \label{time_vs_block_size}
\end{wrapfigure}
We test the performance of our approach for varying subdomain resolutions. The autoencoders in the CoAE-MLSim approach are trained using the data corresponding to the case \ref{cht} with subdomain resolutions of $8^3$, $16^3$ and $32^3$. The trained models are used to solve for PDE solutions for $50$ different unseen heat sources and their results, shown in the table below, are compared with Ansys Fluent on $3$ metrics, Error in maximum temperature (hot spots on chip), $L_{\infty}$ error in temperature, Error in heat flux (temperature gradient) on the chip surface. Table below shows results for $3$ randomly chosen cases and Fig. \ref{time_vs_block_size} plots the computational time averaged over all test cases.

\begin{tabular}{| *{13}{c|} }
    \hline
Case ID   & \multicolumn{3}{c|}{$L_{\infty} (|T_{true} - T_{pred}|)$}
            & \multicolumn{3}{c|}{$Err. (T_{max})$}& \multicolumn{3}{c|}{$Avg.\ Error (flux)$}\\
    \hline
      &$16^3$&$32^3$&$8^3$
      &$16^3$&$32^3$&$8^3$
      &$16^3$&$32^3$&$8^3$\\
    \hline
553   &20.36&16.90&16.93& 5.09 & 5.50 &8.35& 1.26 & 0.10&0.65\\
    \hline
555   &14.35&12.89 &19.45& -1.04 & -1.99&-4.32& 0.59 & 0.09&0.64\\
    \hline
574   &17.80&10.20 &20.1& 8.90 & 5.00&12.20& 1.86 & 0.13&1.97\\
    \hline
\end{tabular}

The accuracy is very similar for different subdomain sizes, but the computational time is drastically different. Higher subdomain resolution corresponds to fewer subdomains in the entire domain and hence reduction in computational cost. The reduction in computational time is not linear because the latent vector compression is smaller for larger subdomains. The extent of subdomain resolutions for the CoAE-MLSim approach ranges from $1^3$ to $n^3$, where $n$ is the total number of voxels in any spatial direction. A $1^3$ subdomain resolution means smaller computational speedups, while $n^3$ means a single subdomain, which would provide the highest computational speedup but with loss in accuracy. Hence, the choice of subdomain size depends on the trade-off between speed and accuracy.

\textbf{Stability:}

There is a long standing challenge in the field of numerical simulation to guarantee the stability and convergence of non-linear PDE solvers. However, we believe that the denoising capability of autoencoders \citep{vincent2010stacked, goodfellow2016deep, du2016stacked, bengio2013generalized, ranzato2007efficient} used in our iterative solution methodology presents a unique benefit, irrespective of the choice of initial conditions. In this work, we empirically demonstrate the stability of our approach for case \ref{cht}.

\begin{wrapfigure}{r}{0.35\textwidth}
    \centering
    \vspace{-3em}
    \includegraphics[width=0.35\textwidth]{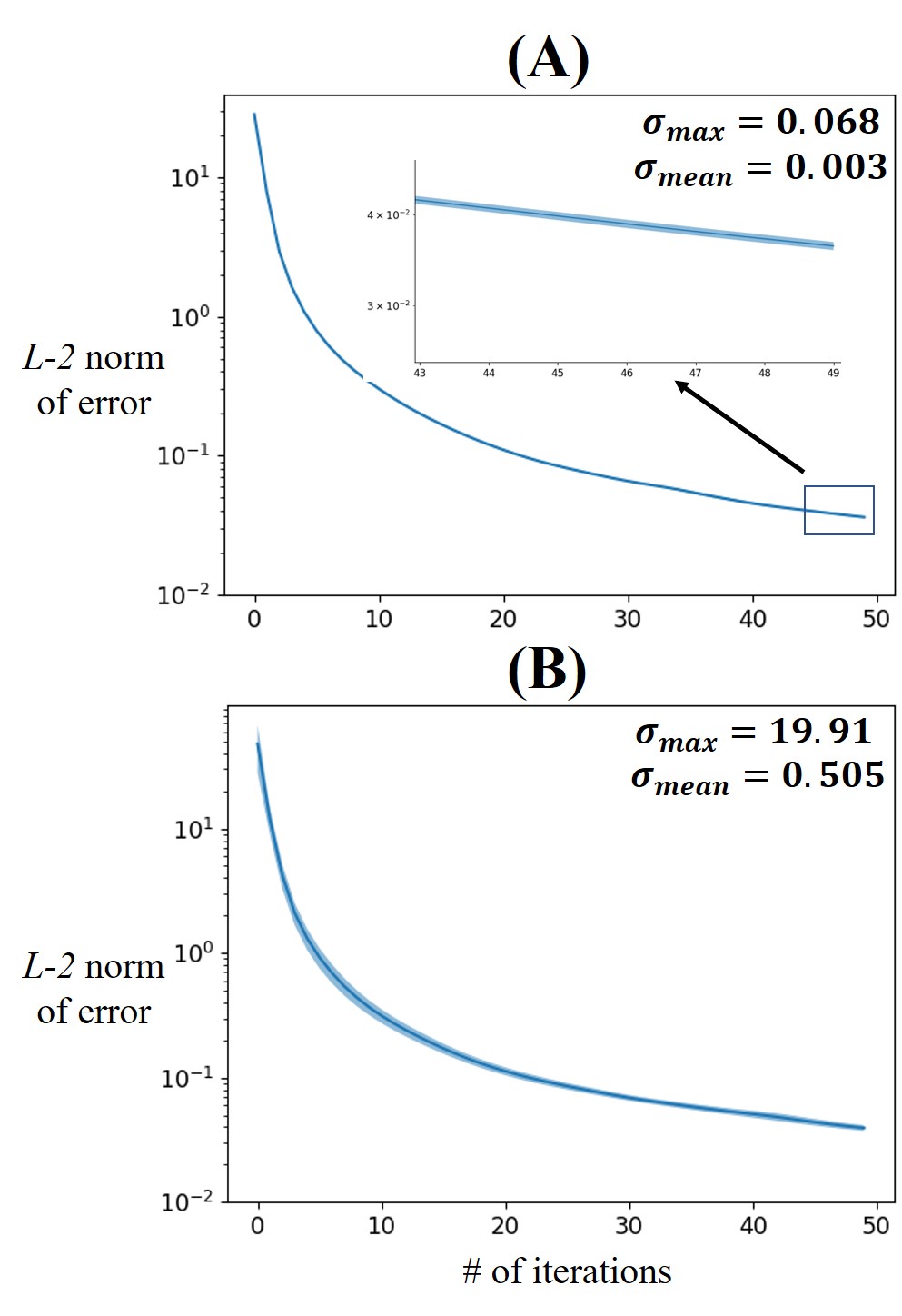}
    \vspace{-2ex}
    \caption{Robustness of CoAE-MLSim approach}
    \label{robustness}
\end{wrapfigure}

In scenario $A$, we randomly sample $25$ initial solutions from a uniform distribution and in scenario $B$ we sample from $10$ different distributions, such as Gumbel, Beta, Logistic etc. The mean convergence trajectory and the standard deviation bounds plotted in Fig. \ref{robustness} show that the $L_2$ norm of the convergence error falls is acceptable for all cases and demonstrates the stability of our approach.

\textbf{Computational speed:}
We observe that the CoAE-MLSim approach is about $40$-$50$x faster in the steady-state cases and about $100$x faster in transient cases as compared to commercial PDE solvers such as Ansys Fluent for the experiments presented in this work. Both CoAE-MLSim approach and Ansys Fluent are solved on a single Xeon CPU with single precision. The mesh resolutions are the same between Fluent and CoAE-MLSim. We expect our approach to scale to multiple CPUs as traditional PDE solvers but single CPU comparisons are provided here for benchmarking in the absence of any parallelism. It is also observed that the acceleration is roughly commensurate to the solution autoencoders compression ratio. Moreover, our algorithm is a python language interpreted code, whereas Ansys Fluent is an optimized, C language pre-compiled code. We expect the C/C++ version of our algorithm to further provide independent speedups (not included in current estimates). 

\textbf{Extent of generalization:} Like other ML-based approaches, the CoAE-MLSim approach operates within certain bounds of generalization with respect to the high dimensional space of PDE conditions. The training samples in this work are generated from limited amount of data for random choices of PDE conditions sampled from a high-dimensional space. For example: the source term in Section \ref{cht} has $4096$ state dimensions. In many cases, the PDE conditions also have sparse representations, which makes generalization tougher. Here, we have demonstrated that our approach can generalize within the space of high-dimensional and sparse PDE conditions without compromising on computational speed and solution accuracy.

\textbf{Coupling with commercial PDE solvers using solution initialization:}

As mentioned earlier, in all the steady-state experiments carried out in this work, all the solution variables are initialized to zero in the CoAE-MLSim solution algorithm. In this experiment, we argue that better initialization of the solution algorithm can result in faster convergence and the better initialization can be obtained by coupling with commercial PDE solvers. To test this hypothesis, we generate $5$ different initial conditions listed below and compare their convergence trajectories with the original case of zero initialization. In each case the solution corresponding to the specified resolution is computed using Ansys Fluent and interpolated on to the $128$x$128$x$128$x used by CoAE-MLSim approach.

\begin{enumerate}[label={(\alph*)}]
    \item Initialization with zero solution (no coupling with Ansys Fluent)
    \item Initialization with coarse resolution Ansys Fluent solution generated on $32$x$32$x$32$ mesh
    \item Initialization with medium resolution Ansys Fluent solution generated on $64$x$64$x$64$ mesh
    \item Initialization with fine resolution Ansys Fluent solution generated on $128$x$128$x$128$ mesh
    \item Initialization with fine resolution Ansys Fluent solution generated on $128$x$128$x$128$ mesh with added random Gaussian noise
\end{enumerate}

\begin{figure}[h!]
  \centering
  \includegraphics[width=0.5\textwidth]{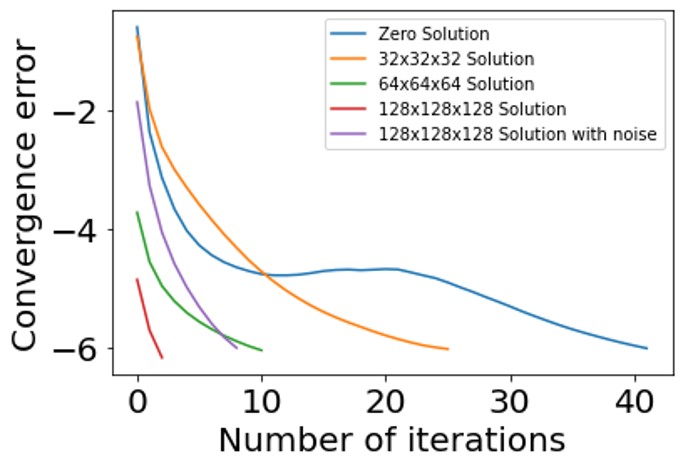}
  \caption{Convergence of CoAE-MLSim solution after coupling with commercial PDE solvers.}
  \label{conv_init1}
\end{figure}

The convergence history comparison on an unseen test geometry is shown in Fig. \ref{conv_init1}. It may be observed that with a zero solution initialization, the CoAE-MLSim iterative solution algorithm takes about $41$ iterations to converge. On the other hand, when initialized with the best possible initial guess, which is the ground truth solution on fine grid resolution, it converges in $2$ iterations. Next, we add a Gaussian random noise of $25$ \% maximum relative to the solution on each computational element. With the added noise, the solution converges in about $8$ iterations. Finally, when initialized with coarse grid solution generated by Ansys Fluent on $32^3$ and $64^3$ resolutions, CoAE-MLSim take $27$ and $11$ iterations respectively. The convergence is still faster than zero solution initialization by $1.5$x and $4$x, respectively. Nonetheless, this experiment shows that the coupling between our iterative algorithm and PDE solvers can result in significant convergence speedup for calculating high fidelity solutions as compared to zero initialization and also demonstrates that our approach can be used as an accelerator to commercial PDE solvers.

\textbf{Comparison with other ML baselines:} We did not find good benchmarks to compare our unsupervised, iterative inferencing algorithm against, however, for the sake of completeness, we have provided comparisons of the experiments from our paper and appendix with ML baselines such as UNet \citep{ronneberger2015u} and FNO \citep{li2020fourier}, wherever available. The results are shown in the table below. All the errors are computed with respect to results computed by a traditional PDE solver. The comparisons are carried over $50$ unseen test cases. In the chip cooling experiment \ref{cht}, the $L_\infty$ error in temperature is compared since that is the most relevant metric to this experiment. For the vortex decay experiment \ref{vortex}, the mean absolute error is averaged over $50$ timesteps. In both cases, the CoAE-MLSim performs better than the baselines and has small training parameter space. In the appendix we present other experiments with comparisons to Ansys Fluent. Since the CoAE-MLSim approach is unsupervised, local and lower-dimensional, it requires lesser amount of data and trainable parameters. For consistency, we have used the same amount of training data to train the ML baselines.

\begin{tabular}{ |p{4cm}||p{1cm}|p{1cm}|p{1cm}|  }
\hline
Experiment& CoAE-MLSim &UNet&FNO\\
\hline
Chip cooling (Temperature)   & 20.36    &117.07&   x\\
Vortex decay &   0.04  & 0.08   &0.09\\
Laplace equations & 0.007 & 0.195& 0.165 \\
\hline
\end{tabular}
\quad
\begin{tabular}{ |p{2.5cm}||p{1.5cm}|  }
\hline
&\# params\\
\hline
CoAE-MLSim&400 $K$\\
UNet &7.418 $M$\\
FNO &465 $K$\\
\hline
\end{tabular}
\label{comparison_table}

\textbf{Handling unstructured meshes:} One of the current limitations of the CoAE-MLSim approach is that it can only handle structured meshes. One of the main components of our approach requires description of PDE solutions on subdomains into lower-dimensional latent vectors. Obtaining such a lower-dimensional representation becomes very challenging when the mesh describing the subdomain is unstructured. Encoding methods used in \citep{remelli2020meshsdf} may be used and will be explored in the future. In our implementation, unstructured meshes are handled by solving on a structured mesh with equivalent or finer resolution to the unstructured mesh followed by bi-linear interpolation. 

\section{Conclusion} \label{conclusion}
In this work, we introduced the CoAE-MLSim approach, which is an unsupervised, low-dimensional and local machine learning approach for solving PDE and generalizing across a wide range of PDE conditions randomly sampled from a high-dimensional distribution. Our approach is inspired from strategies employed in traditional PDE solvers and adopts an iterative inferencing strategy to solve PDE solutions. It consists of several autoencoders that can be easily trained with very few training samples. The proposed approach is demonstrated to predict accurate solutions for a range of PDEs and generalize across sparse and high dimensional PDE conditions.

\textbf{Broader impact and future work:} In this work we aim to combine the ideas developed by traditional PDE solvers with current advancements in machine learning and computational hardware and achieve faster simulations for industrial use cases that can range from weather prediction to drug discovery. Although the proposed ML-model can generalize across a wide range of PDE conditions, but extrapolation to PDE conditions that are significantly different still remains a challenge. However, this work takes a big step towards laying down the framework on how truly generalizable ML-based solvers can be developed. In future, we would like to address these challenges of generalizability and scalability by training autoencoders on random, application agnostic PDE solutions and enforcing PDE-based constraints in the iterative inferencing procedure. In this work, we also demonstrate the potential of a hybrid solver by coupling with traditional PDE solvers. This will be investigated further along with extensions to inverse problems and scale invariance of PDE solutions. Finally, one of the limitations of this work is that it cannot handle unstructured meshes and this will be addressed in a follow-up paper. 



\bibliography{iclr2022_conference}

\begin{thebibliography}{70}
\providecommand{\natexlab}[1]{#1}
\providecommand{\url}[1]{\texttt{#1}}
\expandafter\ifx\csname urlstyle\endcsname\relax
  \providecommand{\doi}[1]{doi: #1}\else
  \providecommand{\doi}{doi: \begingroup \urlstyle{rm}\Url}\fi

\bibitem[Amos \& Kolter(2017)Amos and Kolter]{amos2017optnet}
Brandon Amos and J~Zico Kolter.
\newblock Optnet: Differentiable optimization as a layer in neural networks.
\newblock In \emph{International Conference on Machine Learning}, pp.\
  136--145. PMLR, 2017.

\bibitem[Anandkumar et~al.(2020)Anandkumar, Azizzadenesheli, Bhattacharya,
  Kovachki, Li, Liu, and Stuart]{anandkumar2020neural}
Anima Anandkumar, Kamyar Azizzadenesheli, Kaushik Bhattacharya, Nikola
  Kovachki, Zongyi Li, Burigede Liu, and Andrew Stuart.
\newblock Neural operator: Graph kernel network for partial differential
  equations.
\newblock In \emph{ICLR 2020 Workshop on Integration of Deep Neural Models and
  Differential Equations}, 2020.

\bibitem[Bai et~al.(2019)Bai, Kolter, and Koltun]{bai2019deep}
Shaojie Bai, J~Zico Kolter, and Vladlen Koltun.
\newblock Deep equilibrium models.
\newblock \emph{arXiv preprint arXiv:1909.01377}, 2019.

\bibitem[Bank et~al.(2020)Bank, Koenigstein, and Giryes]{bank2020autoencoders}
Dor Bank, Noam Koenigstein, and Raja Giryes.
\newblock Autoencoders.
\newblock \emph{arXiv preprint arXiv:2003.05991}, 2020.

\bibitem[Bar-Sinai et~al.(2019)Bar-Sinai, Hoyer, Hickey, and
  Brenner]{bar2019learning}
Yohai Bar-Sinai, Stephan Hoyer, Jason Hickey, and Michael~P Brenner.
\newblock Learning data-driven discretizations for partial differential
  equations.
\newblock \emph{Proceedings of the National Academy of Sciences}, 116\penalty0
  (31):\penalty0 15344--15349, 2019.

\bibitem[Battaglia et~al.(2018)Battaglia, Hamrick, Bapst, Sanchez-Gonzalez,
  Zambaldi, Malinowski, Tacchetti, Raposo, Santoro, Faulkner,
  et~al.]{battaglia2018relational}
Peter~W Battaglia, Jessica~B Hamrick, Victor Bapst, Alvaro Sanchez-Gonzalez,
  Vinicius Zambaldi, Mateusz Malinowski, Andrea Tacchetti, David Raposo, Adam
  Santoro, Ryan Faulkner, et~al.
\newblock Relational inductive biases, deep learning, and graph networks.
\newblock \emph{arXiv preprint arXiv:1806.01261}, 2018.

\bibitem[Baydin et~al.(2018)Baydin, Pearlmutter, Radul, and
  Siskind]{baydin2018automatic}
Atilim~Gunes Baydin, Barak~A Pearlmutter, Alexey~Andreyevich Radul, and
  Jeffrey~Mark Siskind.
\newblock Automatic differentiation in machine learning: a survey.
\newblock \emph{Journal of machine learning research}, 18, 2018.

\bibitem[Beatson et~al.(2020)Beatson, Ash, Roeder, Xue, and
  Adams]{beatson2020learning}
Alex Beatson, Jordan Ash, Geoffrey Roeder, Tianju Xue, and Ryan~P Adams.
\newblock Learning composable energy surrogates for pde order reduction.
\newblock \emph{Advances in Neural Information Processing Systems}, 33, 2020.

\bibitem[Bengio et~al.(2013)Bengio, Yao, Alain, and
  Vincent]{bengio2013generalized}
Yoshua Bengio, Li~Yao, Guillaume Alain, and Pascal Vincent.
\newblock Generalized denoising auto-encoders as generative models.
\newblock \emph{arXiv preprint arXiv:1305.6663}, 2013.

\bibitem[Bhattacharya et~al.(2020)Bhattacharya, Hosseini, Kovachki, and
  Stuart]{bhattacharya2020model}
Kaushik Bhattacharya, Bamdad Hosseini, Nikola~B Kovachki, and Andrew~M Stuart.
\newblock Model reduction and neural networks for parametric pdes.
\newblock \emph{arXiv preprint arXiv:2005.03180}, 2020.

\bibitem[Bode et~al.(2021)Bode, Gauding, Lian, Denker, Davidovic, Kleinheinz,
  Jitsev, and Pitsch]{bode2021using}
Mathis Bode, Michael Gauding, Zeyu Lian, Dominik Denker, Marco Davidovic,
  Konstantin Kleinheinz, Jenia Jitsev, and Heinz Pitsch.
\newblock Using physics-informed enhanced super-resolution generative
  adversarial networks for subfilter modeling in turbulent reactive flows.
\newblock \emph{Proceedings of the Combustion Institute}, 38\penalty0
  (2):\penalty0 2617--2625, 2021.

\bibitem[Cai et~al.(2021{\natexlab{a}})Cai, Wang, Fuest, Jeon, Gray, and
  Karniadakis]{cai2021flow}
Shengze Cai, Zhicheng Wang, Frederik Fuest, Young~Jin Jeon, Callum Gray, and
  George~Em Karniadakis.
\newblock Flow over an espresso cup: inferring 3-d velocity and pressure fields
  from tomographic background oriented schlieren via physics-informed neural
  networks.
\newblock \emph{Journal of Fluid Mechanics}, 915, 2021{\natexlab{a}}.

\bibitem[Cai et~al.(2021{\natexlab{b}})Cai, Wang, Wang, Perdikaris, and
  Karniadakis]{cai2021physics}
Shengze Cai, Zhicheng Wang, Sifan Wang, Paris Perdikaris, and George
  Karniadakis.
\newblock Physics-informed neural networks (pinns) for heat transfer problems.
\newblock \emph{Journal of Heat Transfer}, 2021{\natexlab{b}}.

\bibitem[Crutchfield \& McNamara(1987)Crutchfield and
  McNamara]{crutchfield1987equations}
James~P Crutchfield and BS~McNamara.
\newblock Equations of motion from a data series.
\newblock \emph{Complex systems}, 1\penalty0 (417-452):\penalty0 121, 1987.

\bibitem[de~Avila Belbute-Peres et~al.(2018)de~Avila Belbute-Peres, Smith,
  Allen, Tenenbaum, and Kolter]{de2018end}
Filipe de~Avila Belbute-Peres, Kevin Smith, Kelsey Allen, Josh Tenenbaum, and
  J~Zico Kolter.
\newblock End-to-end differentiable physics for learning and control.
\newblock \emph{Advances in neural information processing systems},
  31:\penalty0 7178--7189, 2018.

\bibitem[Du et~al.(2016)Du, Xiong, Wu, Zhang, Zhang, and Tao]{du2016stacked}
Bo~Du, Wei Xiong, Jia Wu, Lefei Zhang, Liangpei Zhang, and Dacheng Tao.
\newblock Stacked convolutional denoising auto-encoders for feature
  representation.
\newblock \emph{IEEE transactions on cybernetics}, 47\penalty0 (4):\penalty0
  1017--1027, 2016.

\bibitem[Dwivedi et~al.(2021)Dwivedi, Parashar, and
  Srinivasan]{dwivedi2021distributed}
Vikas Dwivedi, Nishant Parashar, and Balaji Srinivasan.
\newblock Distributed learning machines for solving forward and inverse
  problems in partial differential equations.
\newblock \emph{Neurocomputing}, 420:\penalty0 299--316, 2021.

\bibitem[Fluent(2015)]{fluent2015ansys}
ANSYS Fluent.
\newblock Ansys fluent.
\newblock \emph{Academic Research. Release}, 14, 2015.

\bibitem[Fukami et~al.(2020)Fukami, Nakamura, and
  Fukagata]{fukami2020convolutional}
Kai Fukami, Taichi Nakamura, and Koji Fukagata.
\newblock Convolutional neural network based hierarchical autoencoder for
  nonlinear mode decomposition of fluid field data.
\newblock \emph{Physics of Fluids}, 32\penalty0 (9):\penalty0 095110, 2020.

\bibitem[Gao et~al.(2021)Gao, Sun, and Wang]{gao2021phygeonet}
Han Gao, Luning Sun, and Jian-Xun Wang.
\newblock Phygeonet: physics-informed geometry-adaptive convolutional neural
  networks for solving parameterized steady-state pdes on irregular domain.
\newblock \emph{Journal of Computational Physics}, 428:\penalty0 110079, 2021.

\bibitem[Gibou et~al.(2018)Gibou, Fedkiw, and Osher]{gibou2018review}
Frederic Gibou, Ronald Fedkiw, and Stanley Osher.
\newblock A review of level-set methods and some recent applications.
\newblock \emph{Journal of Computational Physics}, 353:\penalty0 82--109, 2018.

\bibitem[Goodfellow et~al.(2016)Goodfellow, Bengio, Courville, and
  Bengio]{goodfellow2016deep}
Ian Goodfellow, Yoshua Bengio, Aaron Courville, and Yoshua Bengio.
\newblock \emph{Deep learning}, volume~1.
\newblock MIT press Cambridge, 2016.

\bibitem[Haghighat \& Juanes(2021)Haghighat and Juanes]{haghighat2021sciann}
Ehsan Haghighat and Ruben Juanes.
\newblock Sciann: A keras/tensorflow wrapper for scientific computations and
  physics-informed deep learning using artificial neural networks.
\newblock \emph{Computer Methods in Applied Mechanics and Engineering},
  373:\penalty0 113552, 2021.

\bibitem[Haghighat et~al.(2021)Haghighat, Raissi, Moure, Gomez, and
  Juanes]{haghighat2021physics}
Ehsan Haghighat, Maziar Raissi, Adrian Moure, Hector Gomez, and Ruben Juanes.
\newblock A physics-informed deep learning framework for inversion and
  surrogate modeling in solid mechanics.
\newblock \emph{Computer Methods in Applied Mechanics and Engineering},
  379:\penalty0 113741, 2021.

\bibitem[He \& Pathak(2020)He and Pathak]{he2020unsupervised}
Haiyang He and Jay Pathak.
\newblock An unsupervised learning approach to solving heat equations on chip
  based on auto encoder and image gradient.
\newblock \emph{arXiv preprint arXiv:2007.09684}, 2020.

\bibitem[Hennigh et~al.(2020)Hennigh, Narasimhan, Nabian, Subramaniam,
  Tangsali, Rietmann, Ferrandis, Byeon, Fang, and Choudhry]{hennigh2020nvidia}
Oliver Hennigh, Susheela Narasimhan, Mohammad~Amin Nabian, Akshay Subramaniam,
  Kaustubh Tangsali, Max Rietmann, Jose del~Aguila Ferrandis, Wonmin Byeon,
  Zhiwei Fang, and Sanjay Choudhry.
\newblock Nvidia simnet\^{}$\{$TM$\}$: an ai-accelerated multi-physics
  simulation framework.
\newblock \emph{arXiv preprint arXiv:2012.07938}, 2020.

\bibitem[Holl et~al.(2020)Holl, Koltun, and Thuerey]{holl2020learning}
Philipp Holl, Vladlen Koltun, and Nils Thuerey.
\newblock Learning to control pdes with differentiable physics.
\newblock \emph{arXiv preprint arXiv:2001.07457}, 2020.

\bibitem[Holland et~al.(2019)Holland, Baeder, and Duraisamy]{holland2019field}
Jonathan~R Holland, James~D Baeder, and Karthikeyan Duraisamy.
\newblock Field inversion and machine learning with embedded neural networks:
  Physics-consistent neural network training.
\newblock In \emph{AIAA Aviation 2019 Forum}, pp.\  3200, 2019.

\bibitem[Jin et~al.(2021)Jin, Cai, Li, and Karniadakis]{jin2021nsfnets}
Xiaowei Jin, Shengze Cai, Hui Li, and George~Em Karniadakis.
\newblock Nsfnets (navier-stokes flow nets): Physics-informed neural networks
  for the incompressible navier-stokes equations.
\newblock \emph{Journal of Computational Physics}, 426:\penalty0 109951, 2021.

\bibitem[Kevrekidis et~al.(2003)Kevrekidis, Gear, Hyman, Kevrekidid, Runborg,
  Theodoropoulos, et~al.]{kevrekidis2003equation}
Ioannis~G Kevrekidis, C~William Gear, James~M Hyman, Panagiotis~G Kevrekidid,
  Olof Runborg, Constantinos Theodoropoulos, et~al.
\newblock Equation-free, coarse-grained multiscale computation: Enabling
  mocroscopic simulators to perform system-level analysis.
\newblock \emph{Communications in Mathematical Sciences}, 1\penalty0
  (4):\penalty0 715--762, 2003.

\bibitem[Kharazmi et~al.(2021)Kharazmi, Zhang, and Karniadakis]{kharazmi2021hp}
Ehsan Kharazmi, Zhongqiang Zhang, and George~Em Karniadakis.
\newblock hp-vpinns: Variational physics-informed neural networks with domain
  decomposition.
\newblock \emph{Computer Methods in Applied Mechanics and Engineering},
  374:\penalty0 113547, 2021.

\bibitem[Kim et~al.(2019)Kim, Azevedo, Thuerey, Kim, Gross, and
  Solenthaler]{kim2019deep}
Byungsoo Kim, Vinicius~C Azevedo, Nils Thuerey, Theodore Kim, Markus Gross, and
  Barbara Solenthaler.
\newblock Deep fluids: A generative network for parameterized fluid
  simulations.
\newblock In \emph{Computer Graphics Forum}, volume~38, pp.\  59--70. Wiley
  Online Library, 2019.

\bibitem[Kochkov et~al.(2021)Kochkov, Smith, Alieva, Wang, Brenner, and
  Hoyer]{kochkov2021machine}
Dmitrii Kochkov, Jamie~A Smith, Ayya Alieva, Qing Wang, Michael~P Brenner, and
  Stephan Hoyer.
\newblock Machine learning accelerated computational fluid dynamics.
\newblock \emph{arXiv preprint arXiv:2102.01010}, 2021.

\bibitem[Li et~al.(2021)Li, Hoyer, Pederson, Sun, Cubuk, Riley, Burke,
  et~al.]{li2021kohn}
Li~Li, Stephan Hoyer, Ryan Pederson, Ruoxi Sun, Ekin~D Cubuk, Patrick Riley,
  Kieron Burke, et~al.
\newblock Kohn-sham equations as regularizer: Building prior knowledge into
  machine-learned physics.
\newblock \emph{Physical review letters}, 126\penalty0 (3):\penalty0 036401,
  2021.

\bibitem[Li et~al.(2020{\natexlab{a}})Li, Kovachki, Azizzadenesheli, Liu,
  Bhattacharya, Stuart, and Anandkumar]{li2020fourier}
Zongyi Li, Nikola Kovachki, Kamyar Azizzadenesheli, Burigede Liu, Kaushik
  Bhattacharya, Andrew Stuart, and Anima Anandkumar.
\newblock Fourier neural operator for parametric partial differential
  equations.
\newblock \emph{arXiv preprint arXiv:2010.08895}, 2020{\natexlab{a}}.

\bibitem[Li et~al.(2020{\natexlab{b}})Li, Kovachki, Azizzadenesheli, Liu,
  Bhattacharya, Stuart, and Anandkumar]{li2020multipole}
Zongyi Li, Nikola Kovachki, Kamyar Azizzadenesheli, Burigede Liu, Kaushik
  Bhattacharya, Andrew Stuart, and Anima Anandkumar.
\newblock Multipole graph neural operator for parametric partial differential
  equations.
\newblock \emph{arXiv preprint arXiv:2006.09535}, 2020{\natexlab{b}}.

\bibitem[Lu et~al.(2021{\natexlab{a}})Lu, He, Kasimbeg, Ranade, and
  Pathak]{lu2021one}
Lu~Lu, Haiyang He, Priya Kasimbeg, Rishikesh Ranade, and Jay Pathak.
\newblock One-shot learning for solution operators of partial differential
  equations.
\newblock \emph{arXiv preprint arXiv:2104.05512}, 2021{\natexlab{a}}.

\bibitem[Lu et~al.(2021{\natexlab{b}})Lu, Jin, Pang, Zhang, and
  Karniadakis]{lu2021learning}
Lu~Lu, Pengzhan Jin, Guofei Pang, Zhongqiang Zhang, and George~Em Karniadakis.
\newblock Learning nonlinear operators via deeponet based on the universal
  approximation theorem of operators.
\newblock \emph{Nature Machine Intelligence}, 3\penalty0 (3):\penalty0
  218--229, 2021{\natexlab{b}}.

\bibitem[Lu et~al.(2021{\natexlab{c}})Lu, Meng, Mao, and
  Karniadakis]{lu2021deepxde}
Lu~Lu, Xuhui Meng, Zhiping Mao, and George~Em Karniadakis.
\newblock Deepxde: A deep learning library for solving differential equations.
\newblock \emph{SIAM Review}, 63\penalty0 (1):\penalty0 208--228,
  2021{\natexlab{c}}.

\bibitem[Maleki et~al.(2021)Maleki, Heyse, Ranade, He, Kasimbeg, and
  Pathak]{maleki2021geometry}
Amir Maleki, Jan Heyse, Rishikesh Ranade, Haiyang He, Priya Kasimbeg, and Jay
  Pathak.
\newblock Geometry encoding for numerical simulations.
\newblock \emph{arXiv preprint arXiv:2104.07792}, 2021.

\bibitem[Mao et~al.(2020)Mao, Jagtap, and Karniadakis]{mao2020physics}
Zhiping Mao, Ameya~D Jagtap, and George~Em Karniadakis.
\newblock Physics-informed neural networks for high-speed flows.
\newblock \emph{Computer Methods in Applied Mechanics and Engineering},
  360:\penalty0 112789, 2020.

\bibitem[Maulik et~al.(2020)Maulik, Lusch, and Balaprakash]{maulik2020reduced}
Romit Maulik, Bethany Lusch, and Prasanna Balaprakash.
\newblock Reduced-order modeling of advection-dominated systems with recurrent
  neural networks and convolutional autoencoders.
\newblock \emph{arXiv preprint arXiv:2002.00470}, 2020.

\bibitem[Murata et~al.(2020)Murata, Fukami, and Fukagata]{murata2020nonlinear}
Takaaki Murata, Kai Fukami, and Koji Fukagata.
\newblock Nonlinear mode decomposition with convolutional neural networks for
  fluid dynamics.
\newblock \emph{Journal of Fluid Mechanics}, 882, 2020.

\bibitem[Nabian et~al.(2021)Nabian, Gladstone, and
  Meidani]{nabian2021efficient}
Mohammad~Amin Nabian, Rini~Jasmine Gladstone, and Hadi Meidani.
\newblock Efficient training of physics-informed neural networks via importance
  sampling.
\newblock \emph{Computer-Aided Civil and Infrastructure Engineering}, 2021.

\bibitem[Patel et~al.(2021)Patel, Trask, Wood, and Cyr]{patel2021physics}
Ravi~G Patel, Nathaniel~A Trask, Mitchell~A Wood, and Eric~C Cyr.
\newblock A physics-informed operator regression framework for extracting
  data-driven continuum models.
\newblock \emph{Computer Methods in Applied Mechanics and Engineering},
  373:\penalty0 113500, 2021.

\bibitem[Pfaff et~al.(2020)Pfaff, Fortunato, Sanchez-Gonzalez, and
  Battaglia]{pfaff2020learning}
Tobias Pfaff, Meire Fortunato, Alvaro Sanchez-Gonzalez, and Peter~W Battaglia.
\newblock Learning mesh-based simulation with graph networks.
\newblock \emph{arXiv preprint arXiv:2010.03409}, 2020.

\bibitem[Portwood et~al.(2019)Portwood, Mitra, Ribeiro, Nguyen, Nadiga, Saenz,
  Chertkov, Garg, Anandkumar, Dengel, et~al.]{portwood2019turbulence}
Gavin~D Portwood, Peetak~P Mitra, Mateus~Dias Ribeiro, Tan~Minh Nguyen,
  Balasubramanya~T Nadiga, Juan~A Saenz, Michael Chertkov, Animesh Garg, Anima
  Anandkumar, Andreas Dengel, et~al.
\newblock Turbulence forecasting via neural ode.
\newblock \emph{arXiv preprint arXiv:1911.05180}, 2019.

\bibitem[Qian et~al.(2020)Qian, Kramer, Peherstorfer, and
  Willcox]{qian2020lift}
Elizabeth Qian, Boris Kramer, Benjamin Peherstorfer, and Karen Willcox.
\newblock Lift \& learn: Physics-informed machine learning for large-scale
  nonlinear dynamical systems.
\newblock \emph{Physica D: Nonlinear Phenomena}, 406:\penalty0 132401, 2020.

\bibitem[Raissi \& Karniadakis(2018)Raissi and Karniadakis]{raissi2018hidden}
Maziar Raissi and George~Em Karniadakis.
\newblock Hidden physics models: Machine learning of nonlinear partial
  differential equations.
\newblock \emph{Journal of Computational Physics}, 357:\penalty0 125--141,
  2018.

\bibitem[Raissi et~al.(2019)Raissi, Perdikaris, and
  Karniadakis]{raissi2019physics}
Maziar Raissi, Paris Perdikaris, and George~E Karniadakis.
\newblock Physics-informed neural networks: A deep learning framework for
  solving forward and inverse problems involving nonlinear partial differential
  equations.
\newblock \emph{Journal of Computational Physics}, 378:\penalty0 686--707,
  2019.

\bibitem[Ranade et~al.(2021{\natexlab{a}})Ranade, Hill, He, Maleki, and
  Pathak]{ranade2021latent}
Rishikesh Ranade, Chris Hill, Haiyang He, Amir Maleki, and Jay Pathak.
\newblock A latent space solver for pde generalization.
\newblock \emph{arXiv preprint arXiv:2104.02452}, 2021{\natexlab{a}}.

\bibitem[Ranade et~al.(2021{\natexlab{b}})Ranade, Hill, and
  Pathak]{ranade2021discretizationnet}
Rishikesh Ranade, Chris Hill, and Jay Pathak.
\newblock Discretizationnet: A machine-learning based solver for navier--stokes
  equations using finite volume discretization.
\newblock \emph{Computer Methods in Applied Mechanics and Engineering},
  378:\penalty0 113722, 2021{\natexlab{b}}.

\bibitem[Ranzato et~al.(2007)Ranzato, Poultney, Chopra, LeCun,
  et~al.]{ranzato2007efficient}
Marc Ranzato, Christopher Poultney, Sumit Chopra, Yann LeCun, et~al.
\newblock Efficient learning of sparse representations with an energy-based
  model.
\newblock \emph{Advances in neural information processing systems},
  19:\penalty0 1137, 2007.

\bibitem[Rao et~al.(2020)Rao, Sun, and Liu]{rao2020physics}
Chengping Rao, Hao Sun, and Yang Liu.
\newblock Physics-informed deep learning for incompressible laminar flows.
\newblock \emph{Theoretical and Applied Mechanics Letters}, 10\penalty0
  (3):\penalty0 207--212, 2020.

\bibitem[Remelli et~al.(2020)Remelli, Lukoianov, Richter, Guillard,
  Bagautdinov, Baque, and Fua]{remelli2020meshsdf}
Edoardo Remelli, Artem Lukoianov, Stephan~R Richter, Beno{\^\i}t Guillard,
  Timur Bagautdinov, Pierre Baque, and Pascal Fua.
\newblock Meshsdf: Differentiable iso-surface extraction.
\newblock \emph{arXiv preprint arXiv:2006.03997}, 2020.

\bibitem[Ronneberger et~al.(2015)Ronneberger, Fischer, and
  Brox]{ronneberger2015u}
Olaf Ronneberger, Philipp Fischer, and Thomas Brox.
\newblock U-net: Convolutional networks for biomedical image segmentation.
\newblock In \emph{International Conference on Medical image computing and
  computer-assisted intervention}, pp.\  234--241. Springer, 2015.

\bibitem[Rumelhart et~al.(1985)Rumelhart, Hinton, and
  Williams]{rumelhart1985learning}
David~E Rumelhart, Geoffrey~E Hinton, and Ronald~J Williams.
\newblock Learning internal representations by error propagation.
\newblock Technical report, California Univ San Diego La Jolla Inst for
  Cognitive Science, 1985.

\bibitem[Sanchez-Gonzalez et~al.(2020)Sanchez-Gonzalez, Godwin, Pfaff, Ying,
  Leskovec, and Battaglia]{sanchez2020learning}
Alvaro Sanchez-Gonzalez, Jonathan Godwin, Tobias Pfaff, Rex Ying, Jure
  Leskovec, and Peter Battaglia.
\newblock Learning to simulate complex physics with graph networks.
\newblock In \emph{International Conference on Machine Learning}, pp.\
  8459--8468. PMLR, 2020.

\bibitem[Shukla et~al.(2021)Shukla, Jagtap, and
  Karniadakis]{shukla2021parallel}
Khemraj Shukla, Ameya~D Jagtap, and George~Em Karniadakis.
\newblock Parallel physics-informed neural networks via domain decomposition.
\newblock \emph{arXiv preprint arXiv:2104.10013}, 2021.

\bibitem[Singh et~al.(2017)Singh, Duraisamy, and Zhang]{singh2017augmentation}
Anand~Pratap Singh, Karthikeyan Duraisamy, and Ze~Jia Zhang.
\newblock Augmentation of turbulence models using field inversion and machine
  learning.
\newblock In \emph{55th AIAA Aerospace Sciences Meeting}, pp.\  0993, 2017.

\bibitem[Taghizadeh et~al.(2021)Taghizadeh, Byrne, and
  Wood]{taghizadeh2021explicit}
Ehsan Taghizadeh, Helen~M Byrne, and Brian~D Wood.
\newblock Explicit physics-informed neural networks for non-linear upscaling
  closure: the case of transport in tissues.
\newblock \emph{arXiv preprint arXiv:2104.01476}, 2021.

\bibitem[Toussaint et~al.(2018)Toussaint, Allen, Smith, and
  Tenenbaum]{toussaint2018differentiable}
Marc~A Toussaint, Kelsey~Rebecca Allen, Kevin~A Smith, and Joshua~B Tenenbaum.
\newblock Differentiable physics and stable modes for tool-use and manipulation
  planning.
\newblock 2018.

\bibitem[Um et~al.(2020)Um, Holl, Brand, Thuerey, et~al.]{um2020solver}
Kiwon Um, Philipp Holl, Robert Brand, Nils Thuerey, et~al.
\newblock Solver-in-the-loop: Learning from differentiable physics to interact
  with iterative pde-solvers.
\newblock \emph{arXiv preprint arXiv:2007.00016}, 2020.

\bibitem[Vincent et~al.(2010)Vincent, Larochelle, Lajoie, Bengio, Manzagol, and
  Bottou]{vincent2010stacked}
Pascal Vincent, Hugo Larochelle, Isabelle Lajoie, Yoshua Bengio, Pierre-Antoine
  Manzagol, and L{\'e}on Bottou.
\newblock Stacked denoising autoencoders: Learning useful representations in a
  deep network with a local denoising criterion.
\newblock \emph{Journal of machine learning research}, 11\penalty0 (12), 2010.

\bibitem[Wandel et~al.(2020)Wandel, Weinmann, and Klein]{wandel2020learning}
Nils Wandel, Michael Weinmann, and Reinhard Klein.
\newblock Learning incompressible fluid dynamics from scratch towards fast,
  differentiable fluid models that generalize.
\newblock \emph{arXiv preprint arXiv:2006.08762}, 2020.

\bibitem[Wang et~al.(2021)Wang, Planas, Chandramowlishwaran, and
  Bostanabad]{wang2021train}
Hengjie Wang, Robert Planas, Aparna Chandramowlishwaran, and Ramin Bostanabad.
\newblock Train once and use forever: Solving boundary value problems in unseen
  domains with pre-trained deep learning models.
\newblock \emph{arXiv preprint arXiv:2104.10873}, 2021.

\bibitem[Wang et~al.(2020)Wang, Axelrod, and
  G{\'o}mez-Bombarelli]{wang2020differentiable}
Wujie Wang, Simon Axelrod, and Rafael G{\'o}mez-Bombarelli.
\newblock Differentiable molecular simulations for control and learning.
\newblock \emph{arXiv preprint arXiv:2003.00868}, 2020.

\bibitem[Wiewel et~al.(2020)Wiewel, Kim, Azevedo, Solenthaler, and
  Thuerey]{wiewel2020latent}
Steffen Wiewel, Byungsoo Kim, Vinicius~C Azevedo, Barbara Solenthaler, and Nils
  Thuerey.
\newblock Latent space subdivision: stable and controllable time predictions
  for fluid flow.
\newblock In \emph{Computer Graphics Forum}, volume~39, pp.\  15--25. Wiley
  Online Library, 2020.

\bibitem[Wu et~al.(2018)Wu, Xiao, and Paterson]{wu2018physics}
Jin-Long Wu, Heng Xiao, and Eric Paterson.
\newblock Physics-informed machine learning approach for augmenting turbulence
  models: A comprehensive framework.
\newblock \emph{Physical Review Fluids}, 3\penalty0 (7):\penalty0 074602, 2018.

\bibitem[Xue et~al.(2020)Xue, Beatson, Adriaenssens, and
  Adams]{xue2020amortized}
Tianju Xue, Alex Beatson, Sigrid Adriaenssens, and Ryan Adams.
\newblock Amortized finite element analysis for fast pde-constrained
  optimization.
\newblock In \emph{International Conference on Machine Learning}, pp.\
  10638--10647. PMLR, 2020.

\end{thebibliography}
\bibliographystyle{iclr2022_conference}

\newpage
\appendix
\section{Appendix} \label{appendix}
In the appendix, we provide additional details to support and validate the claims established in the main body of the paper. The appendix section is divided into $3$ sections. In Section \ref{appendix:training}, we provide details related to the network architectures, training mechanics and data generation for the autoencoders used in the CoAE-MLSim approach. In Section \ref{appendix:experiments}, we provide additional details and results for the experiments described in the main paper. Additionally, we have demonstrated the CoAE-MLSim on a different use case. In Section \ref{appendix:reproduce}, we present details for reproducibility and for training the various autoencoders in the CoAE-MLSim approach. 


\subsection{Network Architecture, training mechanics and data generation} \label{appendix:training}

The training portion of the CoAE-MLSim approach proposed in this work corresponds to training of several autoencoders to learn the representations of PDE solutions, conditions, such as geometry, boundary conditions and PDE source terms as well as flux conservation and time integration. We train all the autoencoders with the NVIDIA Tesla V-100 GPU using TensorFlow. The autoencoder training is a one-time cost and is reasonably fast. In this section, we describe details related to the network architectures for the different autoencoders, as well as training mechanics and data generation.

\subsubsection{Geometry autoencoder} \label{appendix:geometry_encoder}

The geometry autoencoder learns to compress signed distance field (SDF) representations of geometry. Mathematically, the signed distance at any point within the geometry is defined as the normal distance between that point and closest boundary of a object. More specifically, for $x \in \mathrm{R}^{n}$ and object(s) $\Omega \subset \mathrm{R}^{n}$, the signed distance field $\phi(x)$ is defined by:
\[
    \phi(x)  =   
    \begin{cases} 
     + d(x, \partial \Omega) & x \in \Omega \\
     - d(x, \partial \Omega) & x \notin \Omega 
   \end{cases}  .
\]
where, $\phi(x)$ is the signed distance field for $x \in \mathrm{R}^{n}$ and objects $\Omega \subset \mathrm{R}^{n}$ \cite{gibou2018review}. \citet{maleki2021geometry} use the same representation of geometry to successfully demonstrate the encoding of geometries.
\begin{figure}[h]
  \centering
  \includegraphics[width=\textwidth]{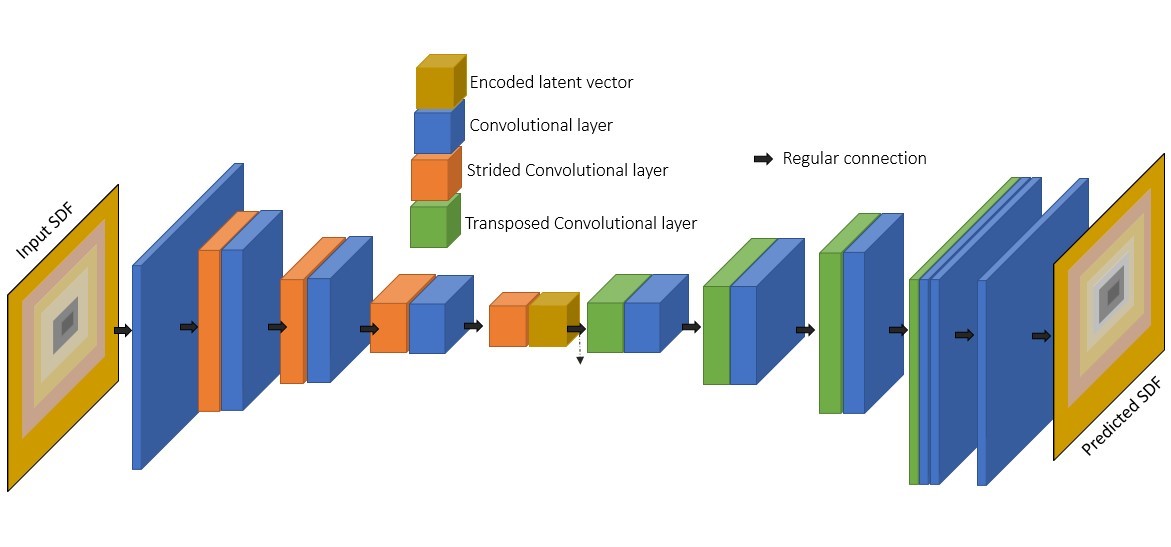}
  \caption{Network architecture for geometry autoencoder (\citet{maleki2021geometry})}
  \label{geometry_arch}
\end{figure}
In this work, we use a CNN-based geometry encoder to encode SDF representation of geometries for relevant use cases. The architecture of this network is shown in Figure \ref{geometry_arch}. The geometry autoencoder has a CNN-based encoder-decoder structure. The encoder compresses the SDF representation to a latent vector, $\eta_g$ and the decoder reconstructs the SDF representation. In the context of CoAE-MLSim approach, a trained geometry encoder is used to represent SDFs on local subdomains with latent vectors. In Section \ref{appendix:geo}, we present results to demonstrate the generalizability of the autoencoder to encode and decode unseen geometries on subdomains.

\textbf{Training data and mechanics:} The training data pertaining to the geometry autoencoder is constructed from random geometries. Arbitrary geometries on entire computational domains are generated and their SDFs are computed. The computational domain is divided into subdomains and the parts of geometry SDFs associated to every subdomains is used as part of the training samples. Since the training is carried out on subdomains, representations of complicated and arbitrary geometries can be learnt accurately. Moreover, we require only $200$-$400$ geometries on entire computational domains to train the autoencoder on subdomains. The autoencoder is trained until a Mean Squared Error (MSE) of $1e^{-6}$ or Mean Absolute Error (MAE) of $1e^{-3}$ is achieved on a validation set. The latent vector length is of the lowest possible size that can result in training errors below these specified thresholds and can be determined through experimentation. Geometry encoders used in all experiments described in this work are trained as mentioned above. 

\subsubsection{PDE source term autoencoder} \label{appendix:source_encoder}

The PDE source term autoencoder learns to compress the spatial distributions of source terms on each subdomain of the computational domain into a latent vector, $\eta_s$. The source term autoencoder uses the same architecture as in Figure \ref{geometry_arch}, except that the inputs and outputs are the source term distributions. 

\textbf{Training data and mechanics:} The training data for source terms is generated on entire computational domains by sampling from a Gaussian mixture model, where the number of Gaussian's, mean and variance of Gaussian's are arbitrary and span over orders of magnitude. The source terms are divided into subdomains, which are used as training samples for the autoencoder. In this case, we generate about $200$-$400$ such source term distributions and train until the MSE or MAE of the validation set drop below their respective thresholds, $1e^{-6}$ and $1e^{-3}$. The latent vector length is chosen such that the training errors are below the specified thresholds. Source term encoders used in relevant experiments described in this work are trained as mentioned above. 

\subsubsection{Representation of boundary conditions} \label{appendix:boundary_encoder}

\begin{figure}[h]
  \centering
  \includegraphics[width=0.75\textwidth]{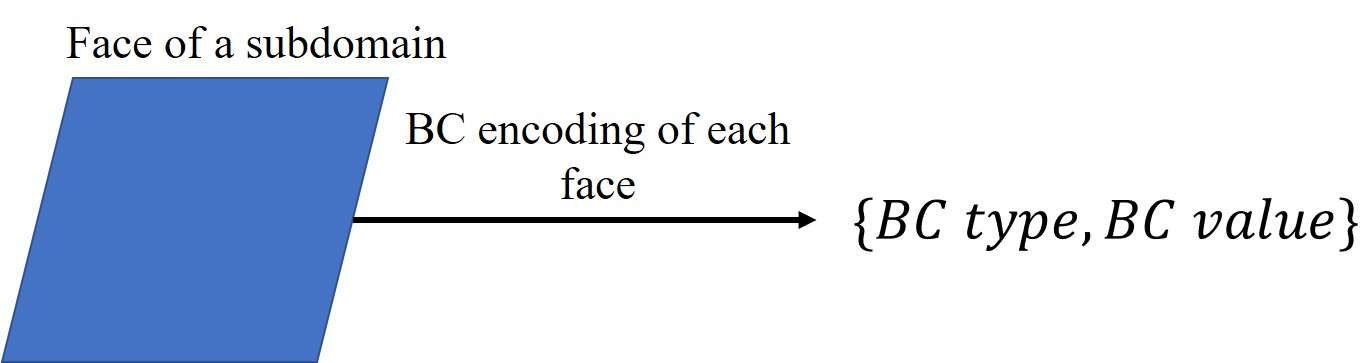}
  \caption{Strategy for BC encoding}
  \label{bc_encode}
\end{figure}

The boundary condition encoders can be learnt using the same autoencoders described in Figure \ref{geometry_arch}. However, in this work we design a manual encoding strategy to establish the latent vectors for boundary conditions. This is because, the choice of boundary conditions in numerical simulations considered in this paper is very limited. The boundary condition encoding strategy is described in Figure \ref{bc_encode}.
On each face of a subdomain, the boundary condition encoding, $\eta_b$, is a vector of size $2$. The first element of this vector indicates the type of boundary condition, followed by the boundary condition value. In this work, $3$ types of boundary conditions are considered, namely Dirichlet, Neumann and Open boundary condition and are indexed as $0, 1, 2$, respectively. Open boundary type refers to faces that do not have any boundary condition imposed on them and is suitable for faces of interior subdomains. For example, a subdomain placed on the inlet boundary, such that the left boundary of the subdomain aligns with the inlet boundary of the entire computational domain, will have a left face encoding of $0, 0.5$, where $0$ refers to Dirichlet boundary and $0.5$ is the BC value. As we move to applications in structural mechanics in future, new methods for encoding boundaries will be introduced.

\subsubsection{PDE solution autoencoder} \label{appendix:solution_encoder}

\begin{figure}[h]
  \centering
  \includegraphics[width=\textwidth]{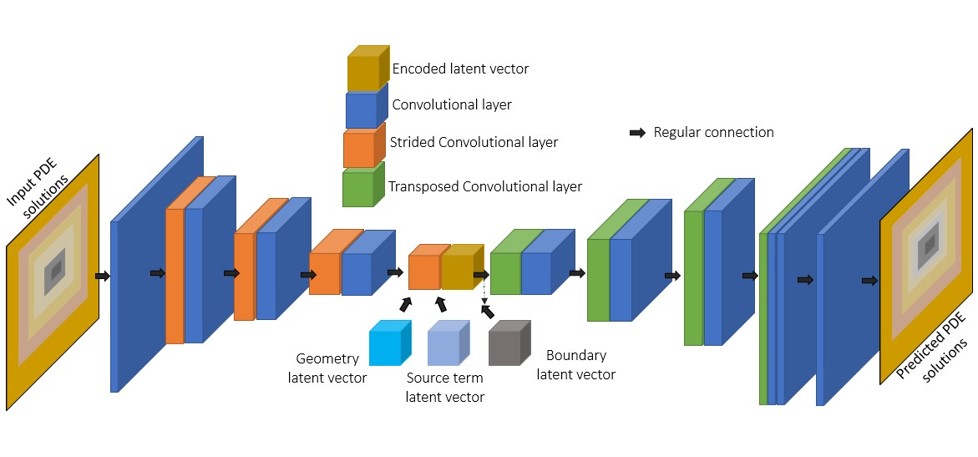}
  \caption{PDE solution autoencoder}
  \label{solution_autoencoder}
\end{figure}

Figure \ref{solution_autoencoder} shows the network architecture used for encoding the solutions, $\eta$, of all PDE solution variables on subdomains. It may be observed that the PDE solution autoencoders are also conditioned on the geometry, source term and boundary latent vectors, that are associated to the subdomains. Since, the PDE solutions are dependent and unique to PDE conditions, establishing this explicit dependency in the autoencoder improves robustness. Additionally, the CoAE-MLSim apprach solves the PDE solution in the latent space, and hence, the idea of conditioning at the bottleneck layer improves solution predictions near geometry and boundaries, especially when the solution latent vector prediction has minor deviations. Each solution variable in the system of PDEs can be trained with a different autoencoder to determine a latent vector which is independent of the latent vector of other solution variables. This strategy of decoupling has shown to increase the accuracy of the solution autoencoders. The specific parameters used in the network architecture can vary based on the size of each subdomain, complexity of physics and extent of compression achieved. 

\textbf{Training data and mechanics:} The PDE solution variables are very specific to the PDEs being solved and the engineering application being modeled. As a result, a different PDE solution autoencoder needs to be trained, when the application and the corresponding PDE is changed. Similar to the other autoencoders, the PDE solution autoencoder is trained on subdomains until an MSE of $1e^{-6}$ or an MAE of $1e^{-3}$ is achieved on a validation set. The compression ratio is selected such that the solution latent vector has the smallest possible size and yet satisfies the accuracy up to these tolerances. Finally, the data for training these autoencoders is generated by running CFD simulations on entire computational domains for arbitrary PDE conditions. The generated solutions are divided into subdomains and used for training the PDE solution autoencoder. Since, the learning is local the number of solutions required to be generated are about $100$-$1000$ for different PDE conditions. The solution autoencoders used in all the experiments have been trained with the strategy described above.

\subsubsection{Flux conservation autoencoders} \label{appendix:flux_encoder}

\begin{figure}[h]
  \centering
  \includegraphics[width=0.75\textwidth]{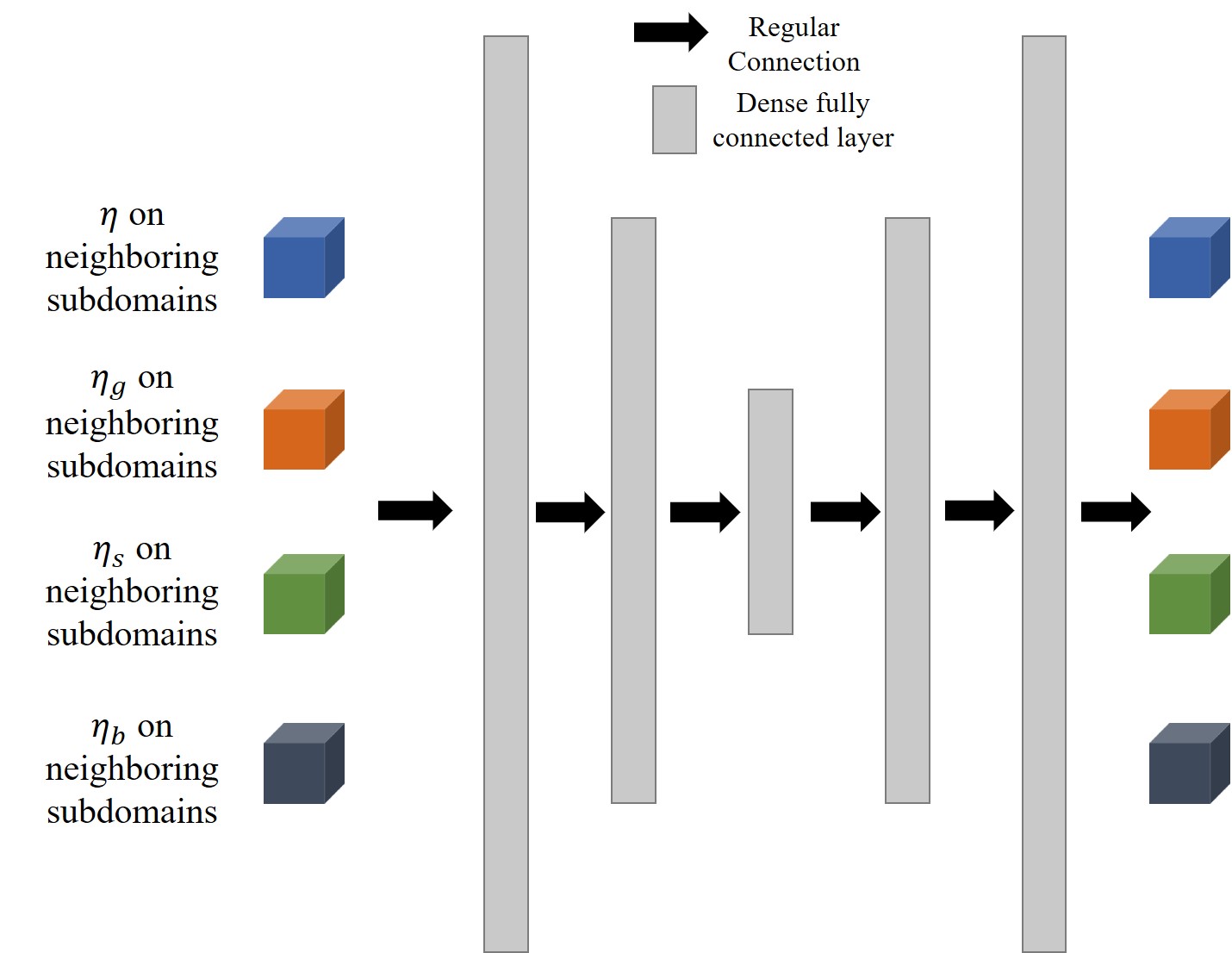}
  \caption{Flux conservation autoencoder}
  \label{flux_arch}
\end{figure}

The flux conservation autoencoder learns the local consistency conditions for a group of neighborhood subdomains in the latent space. Each subdomain is characterized by PDE solutions and conditions and each of these affects the flux conservation autoencoder. As a result, the inputs and outputs of this network are the latent vectors of solution ($\eta$), geometry ($\eta_g$), source term ($\eta_s$) and boundary ($\eta_b$) on a group of neighborhood subdomains. All the solution variables of system of PDEs are stacked together with PDE condition latent vectors and the learnt using the autoencoder architecture shown in Figure \ref{flux_arch}. This autoencoder implicitly learns to represent consistency conditions between neighboring subdomains. Since this autoencoder is only trained on locally consistent subdomains with continuous solutions across intersecting faces, it tries to establish this consistency in neighborhood subdomain solutions for arbitrary inputs. Subdomains at the boundaries may have fewer neighbors and we propose $2$ ways to handle this. Firstly, the information related to the missing neighbors can be substituted with a vector of zeros. This would enable learning of all neighboring subdomain combinations with the same flux conservation network. Conversely, different flux conservation networks can be trained for subdomains with different number of neighbors. For example, a subdomain in the corner will have only $3$ neighbors, while an interior subdomain has $6$ neighbors. In this case, the interior and corner subdomains can be trained separately with different networks. In our experience, both approaches work equally well but we have adopted the approach of zero padding in this work. The specific parameters used in the network architecture can vary based on the size of each subdomain, complexity of physics and extent of compression achieved. 

\textbf{Training data and mechanics:} The training data generated for training the PDE solution autoencoders is used to train these networks as well. The data is pre-processed such that groups of neighboring subdomains are collected together and the solutions and conditions associated with them are encoded using pre-trained autoencoders described in previous sections. This processed data is used for training the flux conservation autoencoder. This autoencoder is trained with an MSE loss and the training stops when the validation loss goes below $1e^{-6}$. The flux conservation autoencoders used in all the experiments have been trained with the strategy described above.

\subsubsection{Time integration autoencoders} \label{appendix:time_encoder}

The network architecture and training data generation and mechanics are very similar to the flux conservation autoencoder. The only difference is in the network architecture, where the time integration autoencoders use latent vectors of PDE solutions at time $t$ and conditions of neighborhood subdomains as the input but only predict the solution latent vectors of all solution variables on the center subdomain of the group of neighborhood subdomains at time $t$+$1$. Each PDE solution variable can be trained with a different time integration autoencoder. The time integration autoencoders used transient PDE related experiments have been trained with the strategy described above.

\subsection{Experiments and additional results} \label{appendix:experiments}
In this section, we provide more results and details of the experiments discussed in the main paper. We have also demonstrated 

\subsubsection{Steady State: Laplace equations} \label{appendix:laplace}

The baseline models  used for comparison in the main paper as well as more information related to our approach is described below:

\textbf{UNet:} The input to this model is a two-channel grid of size 64 x 64. The first channel captures the boundary condition encoding on 2D dimensional grid. There are two boundary conditions chosen for this experiment, Dirichlet and Neumann. On a grid of zeros everywhere, boundaries are coded by replacing zeros with either a 1(Dirichlet) or a 2(Neumann) based on the boundary condition. Similar to the first channel, second channel is also a grid of zeros, and the edges’ zeros are replaced with the magnitude of the boundary condition. The model is trained to output the solution again on a grid of 64 x 64. The UNet has 6 convolutional blocks, 2 at each down-sampled size. The bottleneck size is 8 x 8. The output of the bottleneck is again up-sampled in the usual fashion by concatenating the corresponding down-sampled output. The total number of learnable parameters in UNet baseline is equal to 1.946 million.

\textbf{FNO:} For the Fourier Neural Operator method as well, we have used the same input as in Unet. The FNO model is same as the original implementation in \cite{li2020fourier}. The FNO model has 1.188 million parameters. 

\textbf{FCNN:} In this model, we consider the boundary condition encoding as the input as opposed to a representation of the boundary condition as grid. This network consists of a fully connected network and a convolutional neural network. The boundary condition encoding is first transformed into a vector of size $1024$. This vector is then reshaped into $32$ x $32$ grid. The grid is then passed to convolutional network and the solution is then transformed to $64$ x $64$ grid. The model has $14.8$ million learnable parameters.

For the CoAE-MLSim approach, the computational domain is divided into $64$ subdomains such that each subdomain has a resolution of $8$x$8$ elements. The $64$ dimensional PDE solution is represented by a latent vector of size $7$ using a CNN-based autoencoder. The total number of parameters in the solution and flux autoencoders are $130,000$. The boundary conditions on each subdomain are encoded using the representation provided in Section \ref{appendix:boundary_encoder}. Since, this is a steady state problem, the CoAE-MLSim iterative solution algorithm is initialized with a solution field equal to zero in all test cases.



\subsubsection{Steady State: Electronic chip thermal problem} \label{appendix:cht}
\begin{figure}[h]
  \centering
  \includegraphics[width=0.35\textwidth]{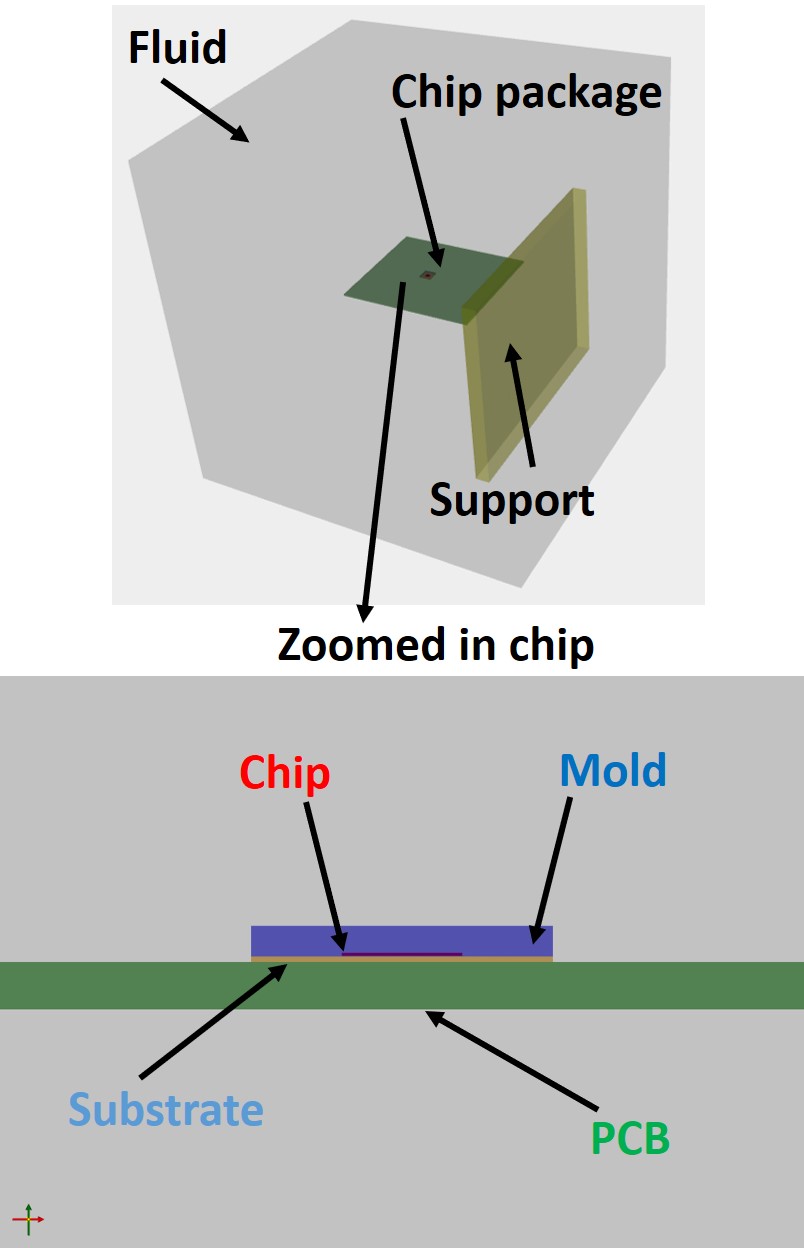}
  \caption{Electronic chip cooling geometry}
  \label{chip_geo}
\end{figure}

\begin{figure}[h]
  \centering
  \includegraphics[width=1.0\textwidth]{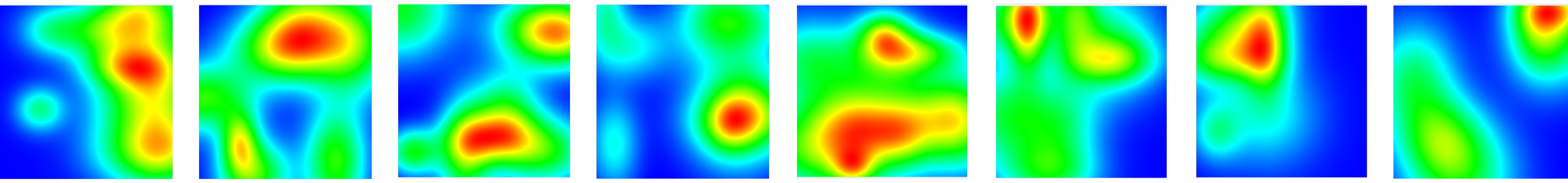}
  \caption{Different power map distributions}
  \label{powermap}
\end{figure}

This is an industrial use case where the domain consists of a chip, which is sandwiched in between an insulated mold. The chip-mold assembly is held by a PCB and the entire geometry is placed inside a fluid domain. The geometry and case setup of the electronic chip cooling case can be observed in Fig. \ref{chip_geo}. The chip is subjected to electric heating and the uncertainty in this process results in random spatial distribution of heat sources on the on the surface of the chip. Fig. \ref{powermap} shows an example of the various distributions of heat sources that the chip might be subjected to due to electrical uncertainty. 

The physics in this problem is natural convection cooling where the power source is responsible for generating heat on the chip, resulting in an increase in chip temperature. The rising temperatures get diffused in to the fluid domain and increase the temperature of air. The air temperature induces velocity which in turn tries to cool the chip. At equilibrium, there is a balance between the chip temperature and velocity generated and both of these quantities reach a steady state. The objective of this problem is to solve for this steady state condition for an arbitrary power source sampled from a Gaussian mixture model distribution, which is extremely high dimensional with up to $25$ Gaussian's, each with a different mean and variance, on a $4096$ dimensional domain. The governing PDEs that represent this problem are shown in Eqs. \ref{eq7}.
\begin{equation}\label{eq7}
\left. \begin{array}{ll}  
\mbox{\textbf{Continuity Equation:} } \quad\quad\quad\quad\quad\quad\quad \quad\quad\displaystyle\nabla . \textbf{v} = 0\quad\quad \quad\quad\quad\quad\quad\quad\quad\quad\\
\mbox{\textbf{Momentum Equation:} } \quad\quad\quad\quad \quad\displaystyle(\textbf{v}.\nabla)\textbf{v} + \nabla \textbf{p} - \frac{1}{Re} \nabla^2 \textbf{\textbf{v}} + \frac{1}{\beta} \Vec{g} \textbf{T} = 0\quad\quad\quad\quad\\[8pt]
\mbox{\textbf{Heat Equation in Solid:} } \quad\quad\quad\quad \quad\quad\displaystyle\nabla .\left( \alpha\nabla\textbf{T}\right)-\color{red}P\color{black} = 0\quad\quad \quad\quad\quad\quad\quad\quad\quad\quad\\[8pt]
\mbox{\textbf{Energy Equation in Fluid:} } \quad\quad\quad\quad\displaystyle(\textbf{v}.\nabla)\textbf{T} - \nabla .\left( \alpha\nabla\textbf{T}\right) = 0\quad\quad\quad\quad\quad\\
 \end{array}\right\}
\end{equation}
where, $v={u_x, u_y, u_z}$ is the velocity field in $x, y, z$, $p$ is pressure, $T$ is temperature, $Re$ and $\alpha$ are flow and thermal properties, $P$ is the heat source term, $\frac{1}{\beta} \Vec{g} T$ is the buoyancy term. $P$ is the spatially varying power source applied on the chip center. The main challenges are in capturing the two-way coupling of velocity and temperature and generalizing over arbitrary spatial distribution of power.

The coupled PDEs with $5$ solution variables, $v={u_x, u_y, u_z}, P, T$ are solved on a fluid and solid domain with loose coupling at the boundaries. The fluid domain is discretized with $128^3$ elements in the domain and the solid domain (chip) is modeled as a 2-D domain with $64^2$ elements as it is very thin in the third spatial dimension.  

\begin{figure}[h!]
  \centering
  \includegraphics[width=0.65\textwidth]{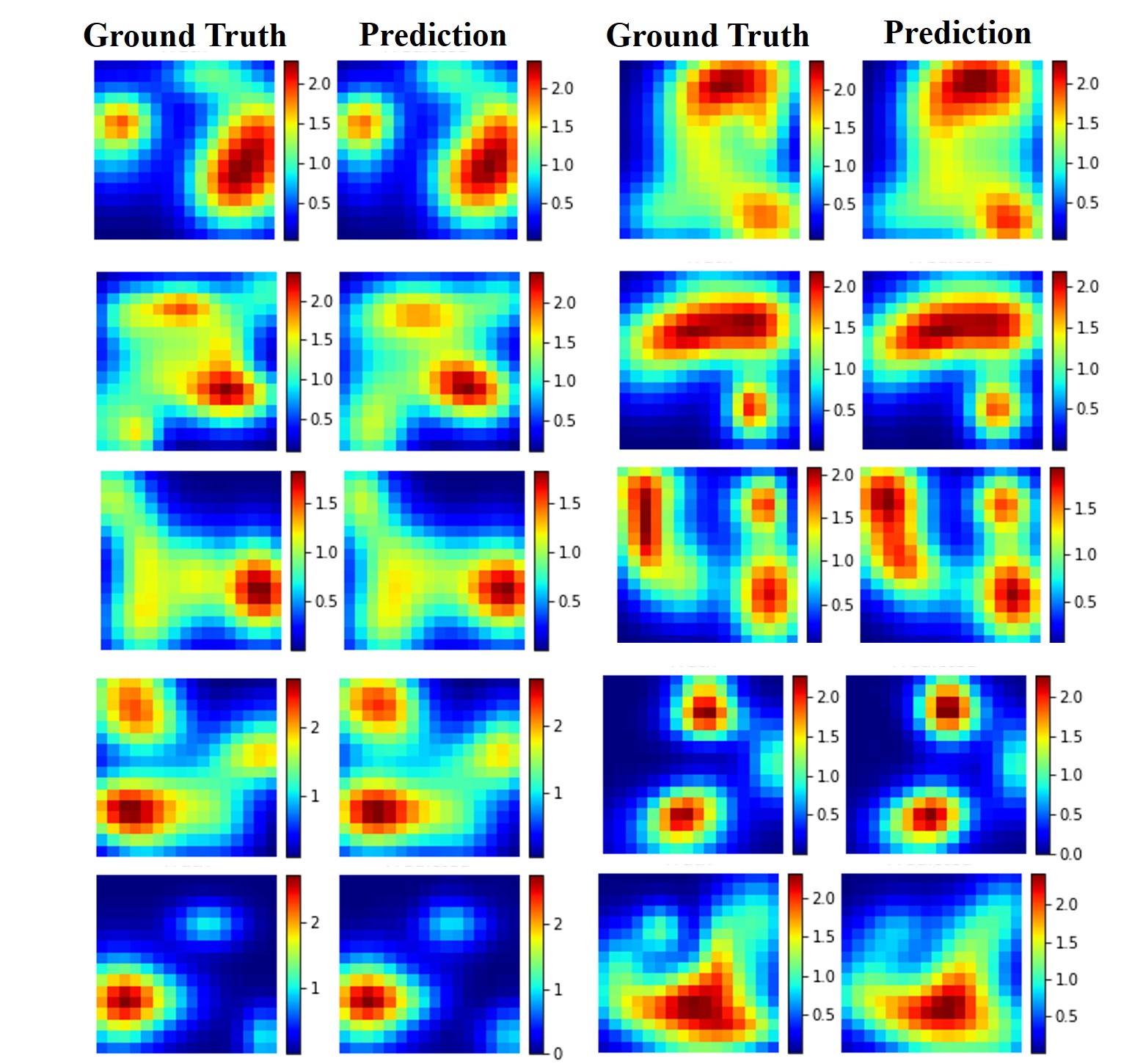}
  \caption{Power source autoencoder for CoAE-MLSim approach}
  \label{source_term_autoencoder}
\end{figure}

The data to train the autoencoders in the CoAE-MLSim approach is generated using Ansys Fluent and corresponds to $300$ PDE solutions. The computational domain is divided into $512$ subdomains, each with $16^3$ computational elements. The solution autoencoders for the $5$ solution variables are trained independently on to establish lower dimensional latent vectors with size $29$ on the subdomain level. The geometry and boundary conditions do not vary and hence an autoencoder is not trained for them in this experiment. The source term autoencoder is trained using randomly generated power source fields. Figure \ref{source_term_autoencoder} shows a few results of the reconstruction capability of the autoencoders. The source term latent vector has a size of $39$.

In the main body of the paper, Section \ref{discussion}, as well as in this section we compare our approach with other baseline ML models such as UNet \citep{ronneberger2015u} for this experiment and here we briefly explain the network architecture used. 

\textbf{UNet:} The architecture of the UNet \citep{ronneberger2015u} is as follows: Since the original dimension of the power map is $64^2$, to construct a $3$D chip power map, zero padding is applied on top and bottom of the $2$D power map, so the input dimension of the Unet is $64^3$. Each module of the contracting path consists of $2$ convolution layers, $1$ max pooling layer and 1 dropout layer. The number of channels changes from $1$ to $48$ in the contracting path and the module is repeated $4$ times. The module in the expansive path consists of $1$ Upsampling layer, $1$ concatenation layer and $2$ convolution layers. Similar to that of the contracting path, the module is repeated $4$ times and the last layer is a convolution layer with $1$ filter. The activation function used is 'Relu', the convolution kernel size is $3$ and the pooling window size is $2$.

\textbf{Additional results:}

Here, we present comparisons between CoAE-MLSim approach and Ansys Fluent for $2$ additional use cases. We compare velocity magnitude and temperature on plane contorus as well as line plots. Also, we compare additional parameters obtained from the simulation such as temperature on PCB-Fluid interface, pressure in the domain, total energy transfer between chip-fluid interface. It may be observed that the results of our approach agree well with a commercial PDE solver and this continues to work for other choices of power sources from the Gaussian mixture model distribution. Since, this is a steady state problem, the CoAE-MLSim iterative solution algorithm is initialized with a solution field equal to zero for all solution variables in all test cases.

\begin{figure}[h!]
  \centering
  \includegraphics[width=0.85\textwidth]{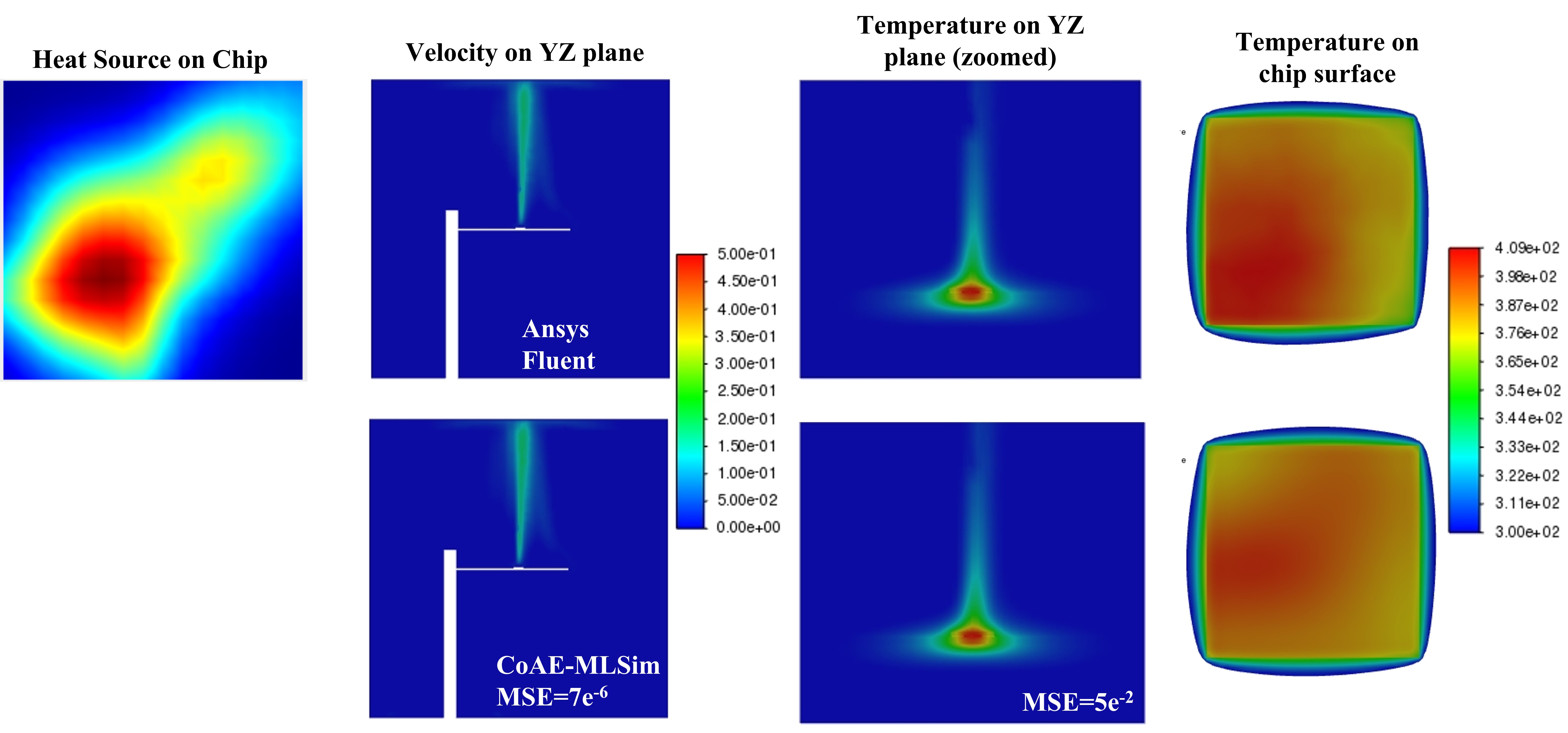}
  \caption{Contour plot comparisons of CoAE-MLSim and Ansys Fluent for test case 1}
  \label{case1_cont_cht_app}
\end{figure}
\begin{figure}[h!]
  \centering
  \includegraphics[width=0.65\textwidth]{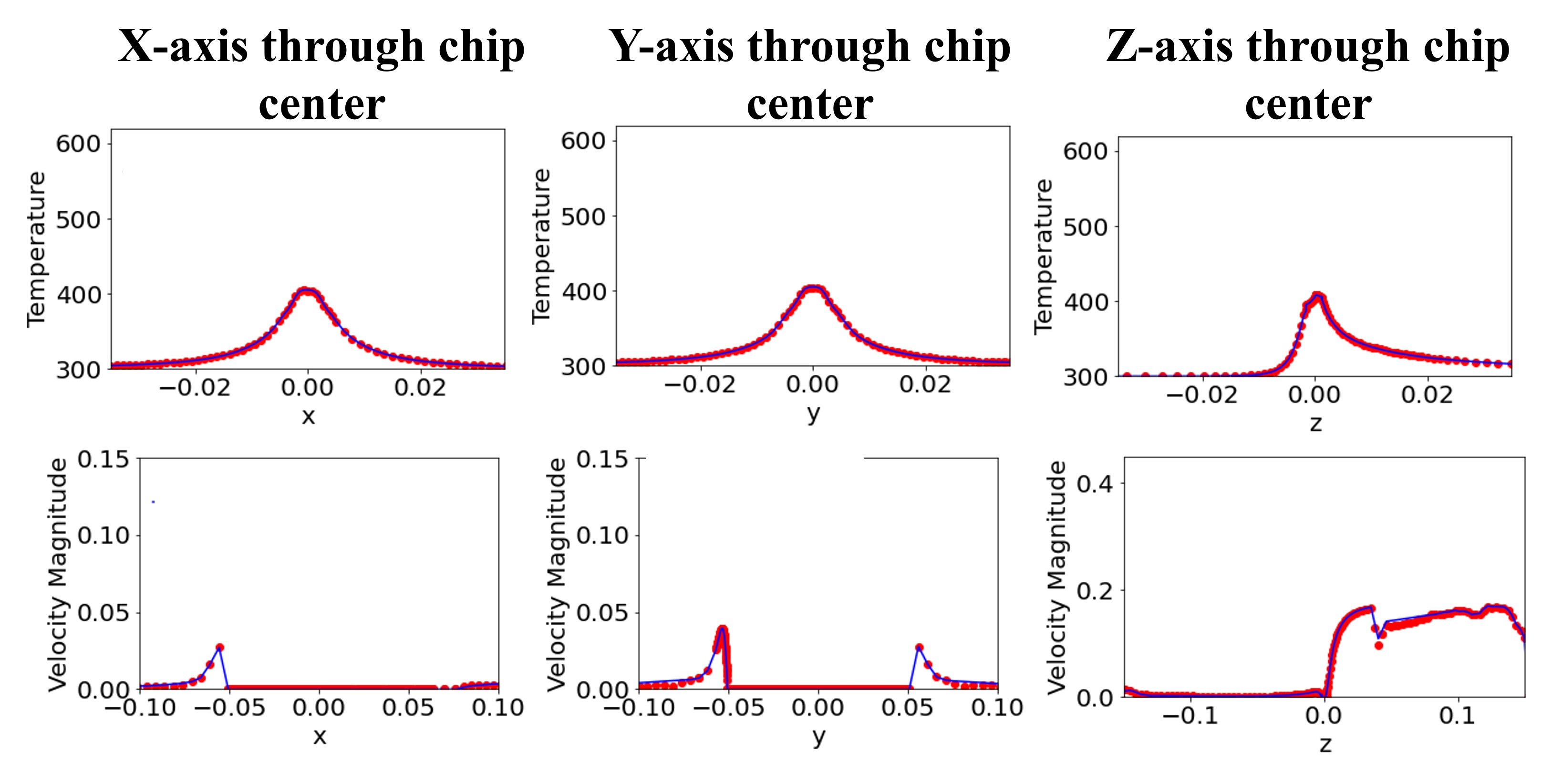}
  \caption{Line plot comparisons of CoAE-MLSim and Ansys Fluent for test case 1 (red: CoAE-MLSim \& blue: Fluent)}
  \label{case1_line_cht_app}
\end{figure}
\begin{figure}[h!]
  \centering
  \includegraphics[width=0.65\textwidth]{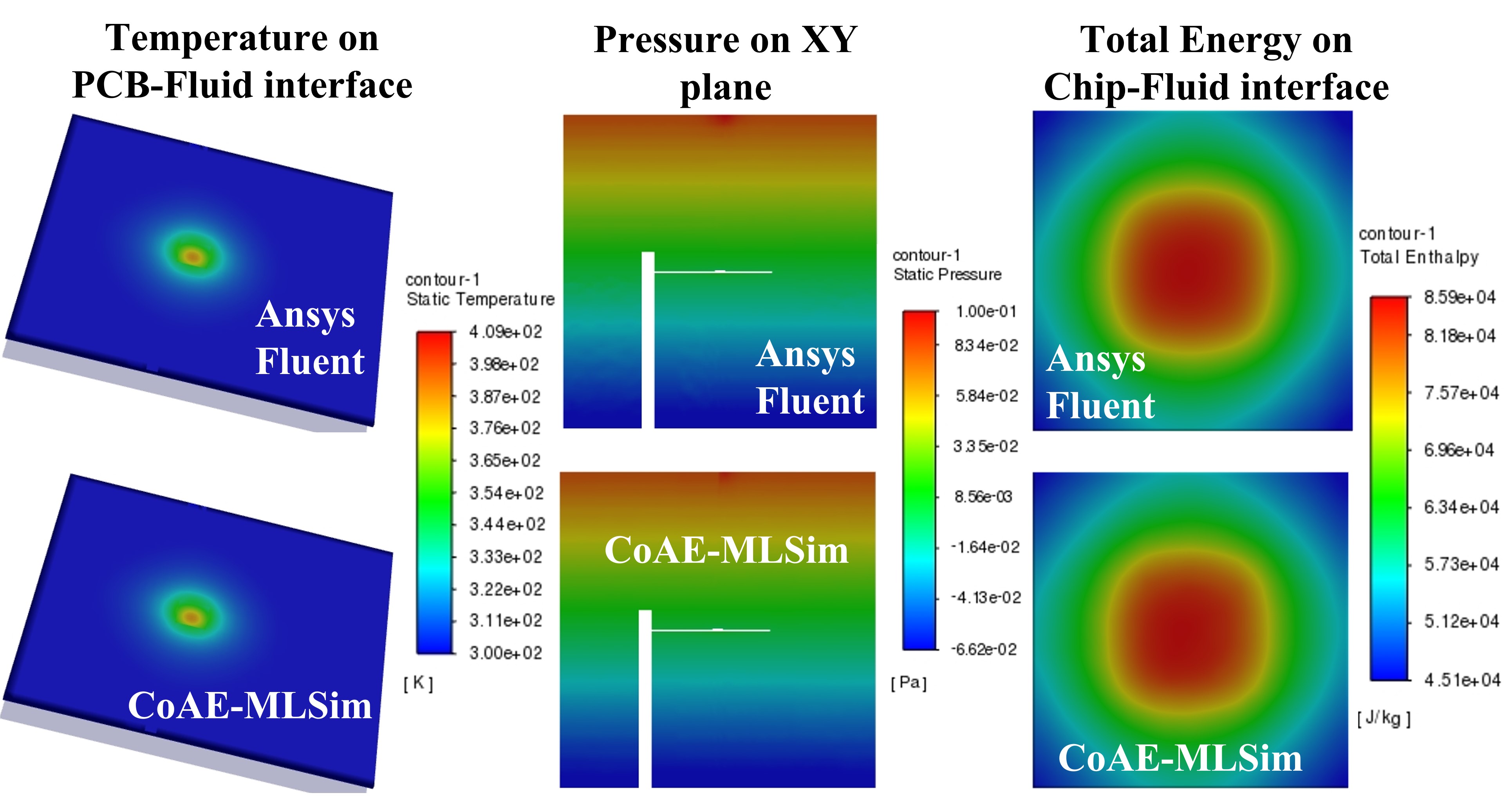}
  \caption{Other parameter contour plot comparisons of CoAE-MLSim and Ansys Fluent for test case 1}
  \label{case1_add_cht_app}
\end{figure}
\begin{figure}[h!]
  \centering
  \includegraphics[width=0.85\textwidth]{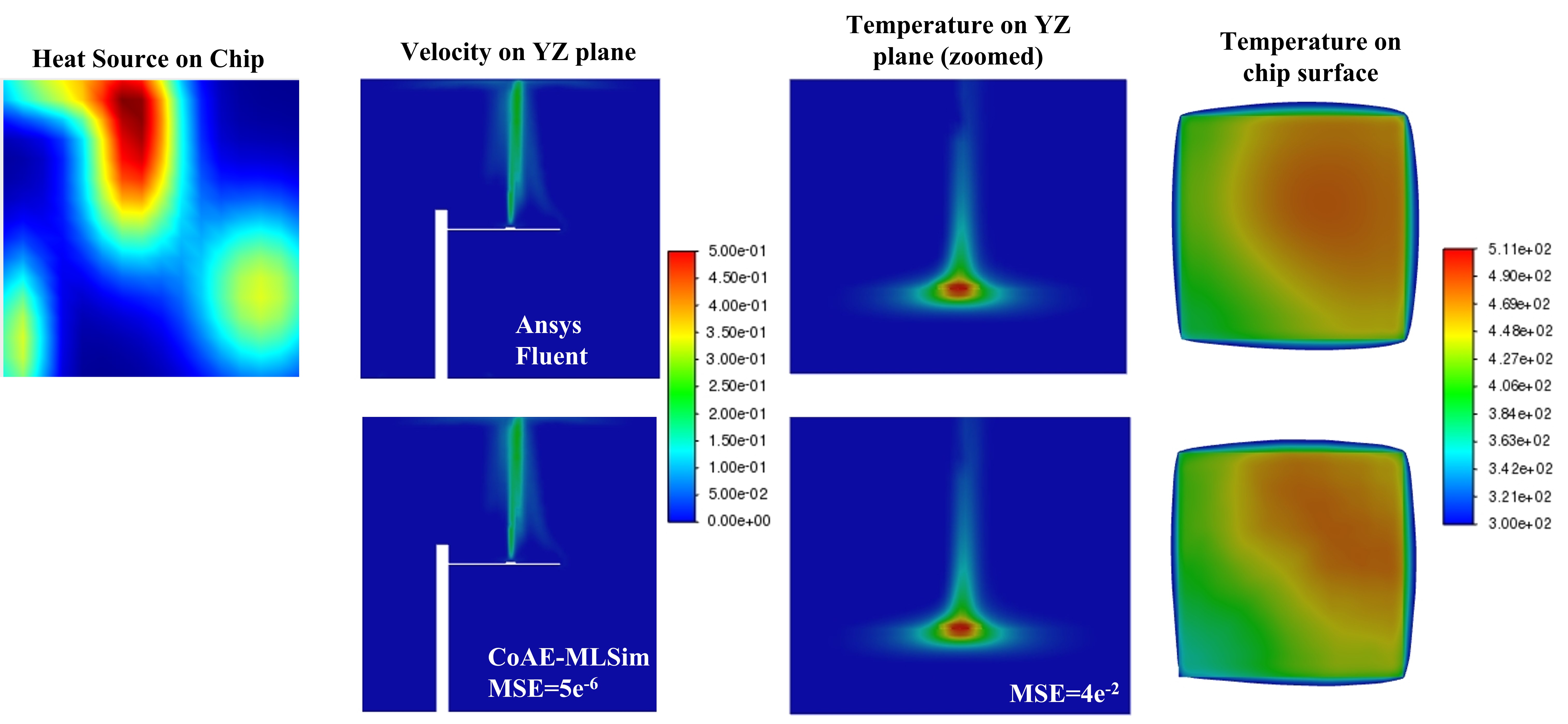}
  \caption{Contour plot comparisons of CoAE-MLSim and Ansys Fluent for test case 2}
  \label{case2_cont_cht_app}
\end{figure}
\begin{figure}[h!]
  \centering
  \includegraphics[width=0.65\textwidth]{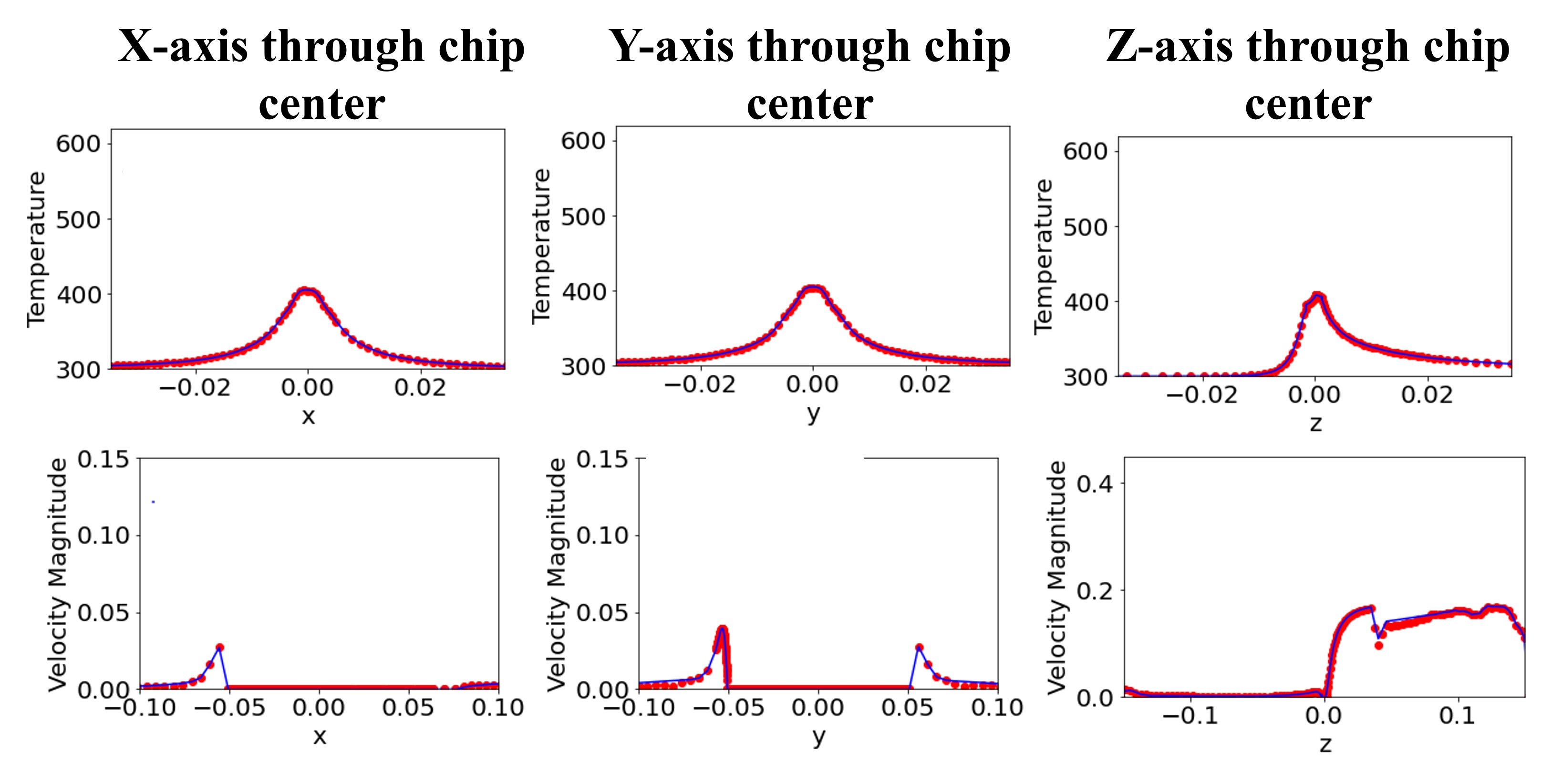}
  \caption{Line plot comparisons of CoAE-MLSim and Ansys Fluent for test case 2 (red: CoAE-MLSim \& blue: Fluent)}
  \label{case2_line_cht_app}
\end{figure}
\begin{figure}[h!]
  \centering
  \includegraphics[width=0.65\textwidth]{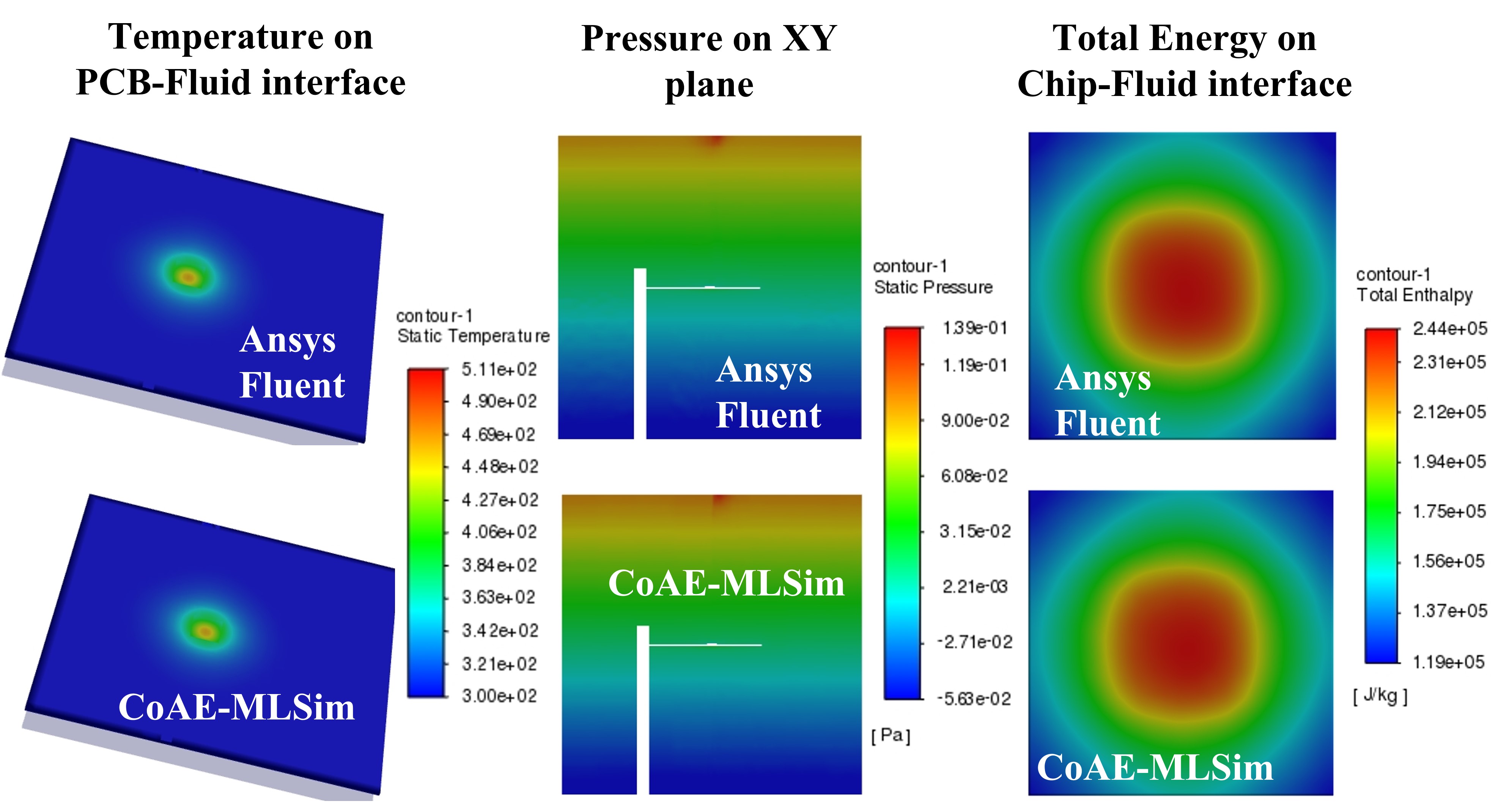}
  \caption{Other parameter contour plot comparisons of CoAE-MLSim and Ansys Fluent for test case 2}
  \label{case2_add_cht_app}
\end{figure}

\bigskip

Finally, we use this experiment to provide further analysis of the CoAE-MLSim approach. The different sub-experiments are listed below:
\begin{enumerate}
    \item Comparison against UNet for various test cases.
    \item Analysis of solution convergence during different iterations of the CoAE-MLSim approach
\end{enumerate}
    
\textbf{Comparison of nearest neighbor, CoAE-MLSim and Unet:}

Since we use very few training samples to train the CoAE-MLSim approach, it is important to demonstrate that our approach is not memorizing and that the physics represented by the experiments is non-trivial. Hence, we investigate this by comparing our approach with a trivial nearest neighbor interpolation and Unet \citep{ronneberger2015u}. We use all methods to solve for $5$ unseen power sources and the results obtained are compared with Ansys Fluent with respect to the metrics discussed below. 
\begin{enumerate}
    \item Error in maximum temperature in computational domain (hot spots on chip), 
    \item $L_{\infty}$ error in temperature in the computational domain,
    \item Error in heat flux (temperature gradient) on the chip surface
\end{enumerate}
These metrics are more suited for engineering simulations and provide a much better measure for evaluating accuracy and generalization than average based measures. The state-of-the-art Unet \citep{ronneberger2015u} is trained on the same number of training samples as used by the CoAE-MLSim approach using the architecture described above. On the other hand, for the nearest-neighbor interpolation we calculate the nearest solution by averaging the solutions obtained from the $3$ closest neighbors (based on Euclidean distance of source terms) in the training set of $600$ simulations, twice more than what is used for training the CoAE-MLSim approach. The results are compared in the table below. 

\begin{center}
\begin{tabular}{| *{10}{c|} }
    \hline
Case ID   & \multicolumn{3}{c|}{$L_{\infty} (|T_{true} - T_{pred}|)$}
            & \multicolumn{3}{c|}{$Error (T_{max})$}& \multicolumn{3}{c|}{$Avg.\ Error (heat\ flux)$}\\
    \hline
      &nearest&CoAE-MLSim&Unet&nearest&CoAE-MLSim&Unet&nearest&CoAE-MLSim&Unet\\
    \hline
553   & 187.45 & 20.36 &117.07 & 186.76 & 5.09 &-117.06 &36.65&1.26&7.41\\
    \hline
554   & 93.76 & 10.50  & 27.24 & -87.97 & 3.40&27.25&17.08&1.45&0.93\\
    \hline
555   & 62.22 & 14.35  & 62.08 & -53.78 & -1.04&-62.08 &9.43&0.59&2.76\\
    \hline
575   & 82.91 & 15.60  & 195.52 & -74.51 & -0.75&-195.51 &14.62&0.80&15.78\\
    \hline
574   & 72.82 & 17.80 & 74.72 & -39.78 & 8.90&-74.72&6.81 & 1.86&3.41\\
    \hline
\end{tabular}
\label{comparison_unet_nn}
\end{center}     

It may be observed that the our approach outperforms the nearest neighbor interpolation by a very large margin on all the metrics as well as the UNet. The UNet is better than the nearest neighbor approach but severely under-performs in comparison to our approach.  

\textbf{Evolution of PDE solution during iterative inferencing:}

\begin{figure}[h!]
  \includegraphics[width=\textwidth]{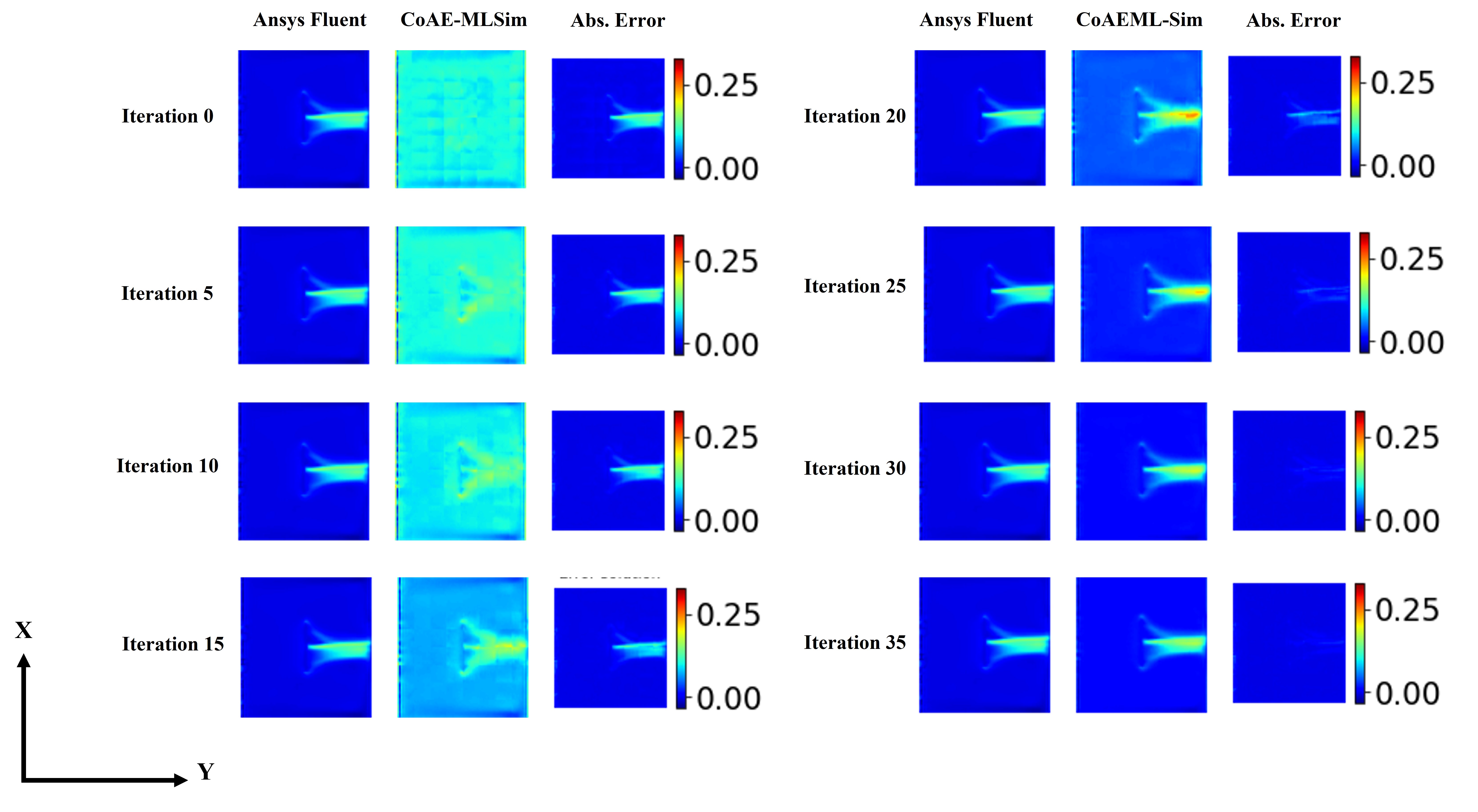}
  \caption{Convergence of CoAE-MLSim solution through the iterative procedure}
  \label{conv_hist}
\end{figure}

Figure \ref{conv_hist} shows the evolution of the CoAE-MLSim solution on a plane cut through the center of the chip and normal to the $z$ direction, at different iterations until convergence. The results are shown for the $y$ component of velocity for test case $1$ presented in Section \ref{cht} and compared with a converged Ansys Fluent solution for the same case. The initial solution provided to the solver is sampled from a uniform random distribution. At iteration $1$,  it may be observed that the flux conservation autoencoder denoises the random signals. In the following iterations, the solution begins to develop based on the source term encoding constraint and finally converges at iteration $35$. The stable progression of the solution points to the robustness and convergence of the CoAE-MLSim.

\subsubsection{Steady State: Flow over arbitrary objects} \label{appendix:geo}

\begin{figure}[h!]
  \centering
  \includegraphics[width=0.7\textwidth]{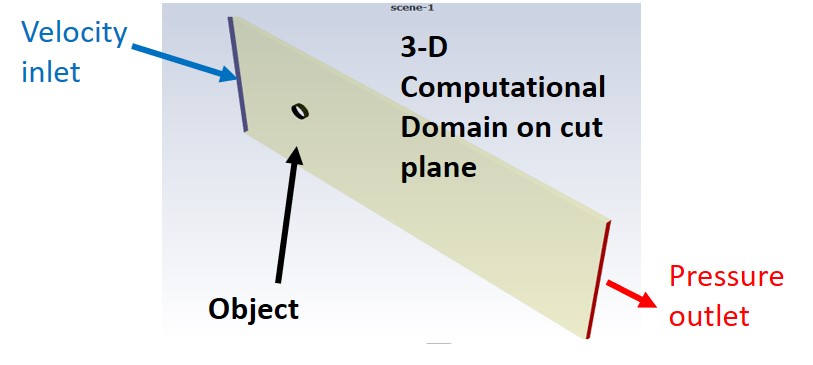}
  \caption{Computational domain}
  \label{appfig:14}
\end{figure}

In this use case, the CoAE-MLSim is demonstrated for generalizing across a wide range of geometry conditions. The geometry of objects are represented with a signed distance field representation and is extremely high-dimensional ($512$-$4096$). The use case consists of a 3-D channel flow over arbitrarily shaped objects as shown in Fig. \ref{appfig:14}. The domain has a velocity inlet specified at $1 m/s$ and a zero pressure outlet boundary conditions on $2$ surfaces, while the rest of the surfaces are walls with no-slip conditions. The shape of the object immersed in flow is arbitrary and the objective is to demonstrate the generalization of the CoAE-MLSim for such geometric variations. 

The computational domain, in this case, is elongated in the $x$ direction, such that it consists of $320$ computational elements in that direction, while the $y$ and $z$ directions have $160$ and $80$ computational elements each. The domain is decomposed into subdomains with $16$ elements in each direction. Boundary conditions, geometry and PDE solutions are encoded on each subdomain using pre-trained encoders. The PDEs solved in this problem consist of $4$ solution variables, include $3$ components of velocity and pressure. The governing equations corresponding to these variables are shown in Eq. \ref{appeq:2}. The subdomain solution of each solution variable is compressed to a latent vector of size $31$. The geometry is represented using signed distance fields, which are compressed to a latent vector of size $29$. Boundary conditions are encoded using the boundary encoder described in Section \ref{boundary_encoder}. There is no source term in this case and hence, it is excluded from the flux conservation autoencoder. The training data in this case corresponds to $300$-$400$ solutions generated for arbitrary geometries using Ansys Fluent. Since, this is a steady state problem, the CoAE-MLSim iterative solution algorithm is initialized with a solution field equal to zero for all solution variables in all test cases.

\begin{equation}\label{appeq:2}
\left. \begin{array}{ll}  
\mbox{\textbf{Continuity Equation:} } \quad\quad\quad\quad\quad\quad\quad \quad\quad\displaystyle\nabla . \textbf{v} = 0\quad\quad \quad\quad\quad\quad\quad\quad\quad\quad\\
\mbox{\textbf{Momentum Equation:} } \quad\quad\quad\quad \quad\displaystyle(\textbf{v}.\nabla)\textbf{v} + \nabla \textbf{p} - \frac{1}{Re} \nabla^2 \textbf{\textbf{v}} = 0\quad\quad\quad\quad\\[8pt]
 \end{array}\right\}
\end{equation}
where, $v={u_x, u_y, u_z}$ is the velocity field in $x, y, z$, $p$ is pressure, $T$ is temperature, $Re$ is Reynolds number representing flow properties.

\begin{figure}[h!]
 \centering
  \includegraphics[width=\textwidth]{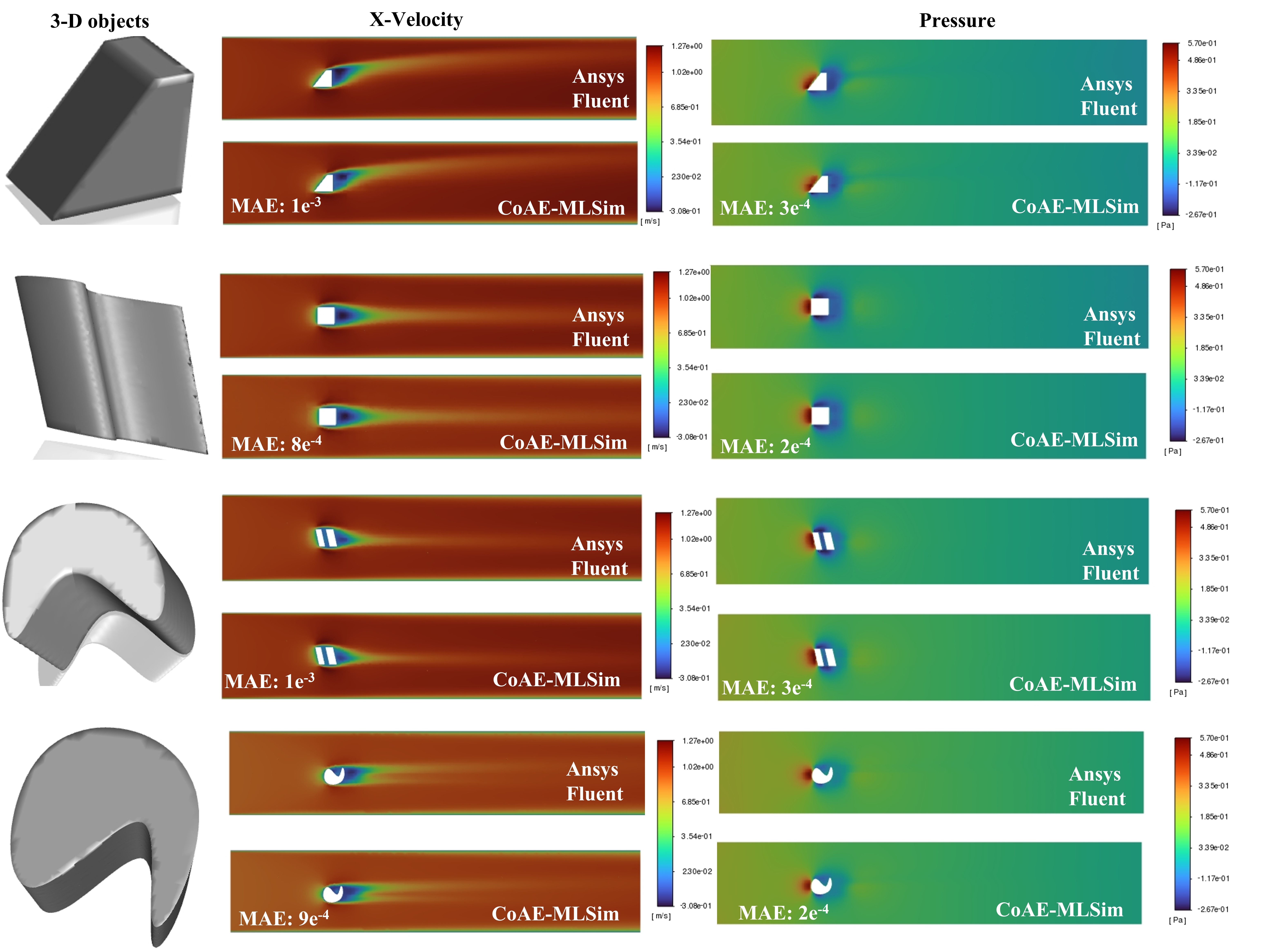}
  \caption{CoAE-MLSim vs Ansys Fluent comparisons for different objects}
  \label{appfig:15}
\end{figure}

\textbf{Additional examples}
Figure \ref{appfig:15} shows comparisons of CoAE-MLSim with Ansys Fluent for $4$ uneen objects in addition to the example shown in the main paper. The solver generalizes well and the errors fall within an acceptable range.

\textbf{Geometry encoder performance}

\begin{figure}[h!]
 \centering
  \includegraphics[width=0.75\textwidth]{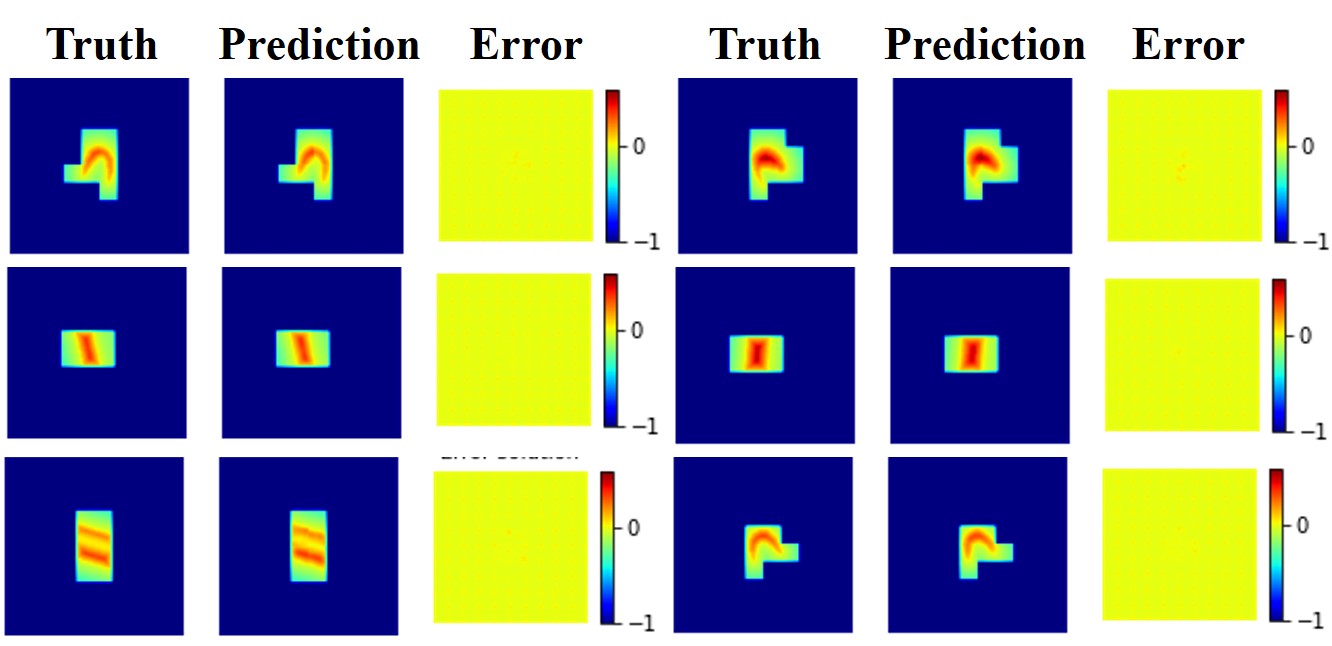}
  \caption{Geometry encoder generalization}
  \label{appfig:15a}
\end{figure}

Figure \ref{appfig:15a} shows the comparison between the true signed distance field and the reconstructed field from the geometry autoencoder. In this example, the geometry autoencoder is evaluated on each subdomain, but all subdomains are reassembled using their connectivity information to obtain the SDF on the entire computational domain. The contour plots in Figure \ref{appfig:15a} are on planes cut through the center of the geometry along $x$ direction. It may be observed that the autoencoder reconstructs agree well with the ground truth for unseen cases. The mean absolute error in all the cases presented here is less than $1e^{-3}$.

\subsubsection{Transient: Vortex decay flow} \label{appendix:decay}

This is a similar use case from \cite{li2020fourier} where a 2-d Navier-Stokes equation for a viscous, incompressible fluid in vorticity form on
the unit torus. The PDE and case details corresponding to this may be found in \cite{li2020fourier}. In the main paper, we provided a single result for an unseen initial condition when integrated in time. Here, we will provide visual comparisons of additional results and additional details about the CoAE-MLSim approach used for this problem.

As described earlier, the mesh resolution considered in this experiment consists of $64$x$64$ computational elements. The dataset considered in this experiment corresponds to a viscosity of $1e^{-3}$ and contains transient evolution of $5000$ initial solution fields over $50$ timsesteps. The autoencoders of the CoAE-MLSim approach are trained with data corresponding to only $500$ initial conditions and first $25$ timesteps. The computational domain is divided into $64$ subdomains of $8^2$ resolution. The solution autoencoder encodes vorticity on each subdomain into a latent vector of size $11$. The flux conservation and time integration autoencoder are trained using the same data. The autoencoder networks combine to have a total of $400, 000$ parameters. The time integration autoencoder is designed such that the input contains solution latent vectors on neighboring subdomains of $10$ previous time steps to predict the latent vectors of the next time step. As a result, all the test runs start from the $10^{th}$ timestep. During each timestep integration during the solution inference, the flux conservation autoencoder is evaluated until an iterative convergence is achieved, similar to the steady state version of the CoAE-MLSim approach and this convergence is achieved to a specified tolerance of $1e^{-6}$ in about $2$-$4$ iterations. As a result, the total evaluation time might be slower than other ML-approaches but is still faster than commercial PDE solvers by around $100$x and moreover, it provides for a more stable and accurate transient dynamics.

In the main body of the paper, Section \ref{discussion}, as well as in this section we compare our approach with other baseline ML models such as UNet \citep{ronneberger2015u} and Fourier Opertor Net (FNO) for this experiment and here we briefly explain the network architectures used. 

\textbf{UNet:} Our UNet architecture includes 18 convolutional layers with channel sizes of 10 (input), 64, 64, 128, 128, 128, 128, 256, 256, 256, 512, 384, 384, 256, 256, 192, 192, 128, 64, 1 (output)  with 9 before the bottleneck (encoder) and 9 after the bottleneck (decoder). Skip connections were used to connect layers before and after the bottleneck. In the encoder, the spatial resolution was reduced using four max-pooling layers with kernel size 2. In the decoder the original spatial resolution was recovered using four bilinear upsampling layers with kernel size 2. The total number of learnable parameters is 7.418M.

\textbf{FNO:} Our FNO architecture was that of the original implementation \citep{li2020fourier}. The total number of learnable parameters is 465k.

Next, we compare results for $3$ unseen initial conditions. It may be observed that the CoAE-MLSim predictions match well with the ground truth data and the error accumulation is acceptable, especially in the extrapolation range. It may be observed that our method outperforms FNO and UNet in terms of error accumulation.

\begin{figure}[h]
  \centering
  \includegraphics[width=0.8\textwidth]{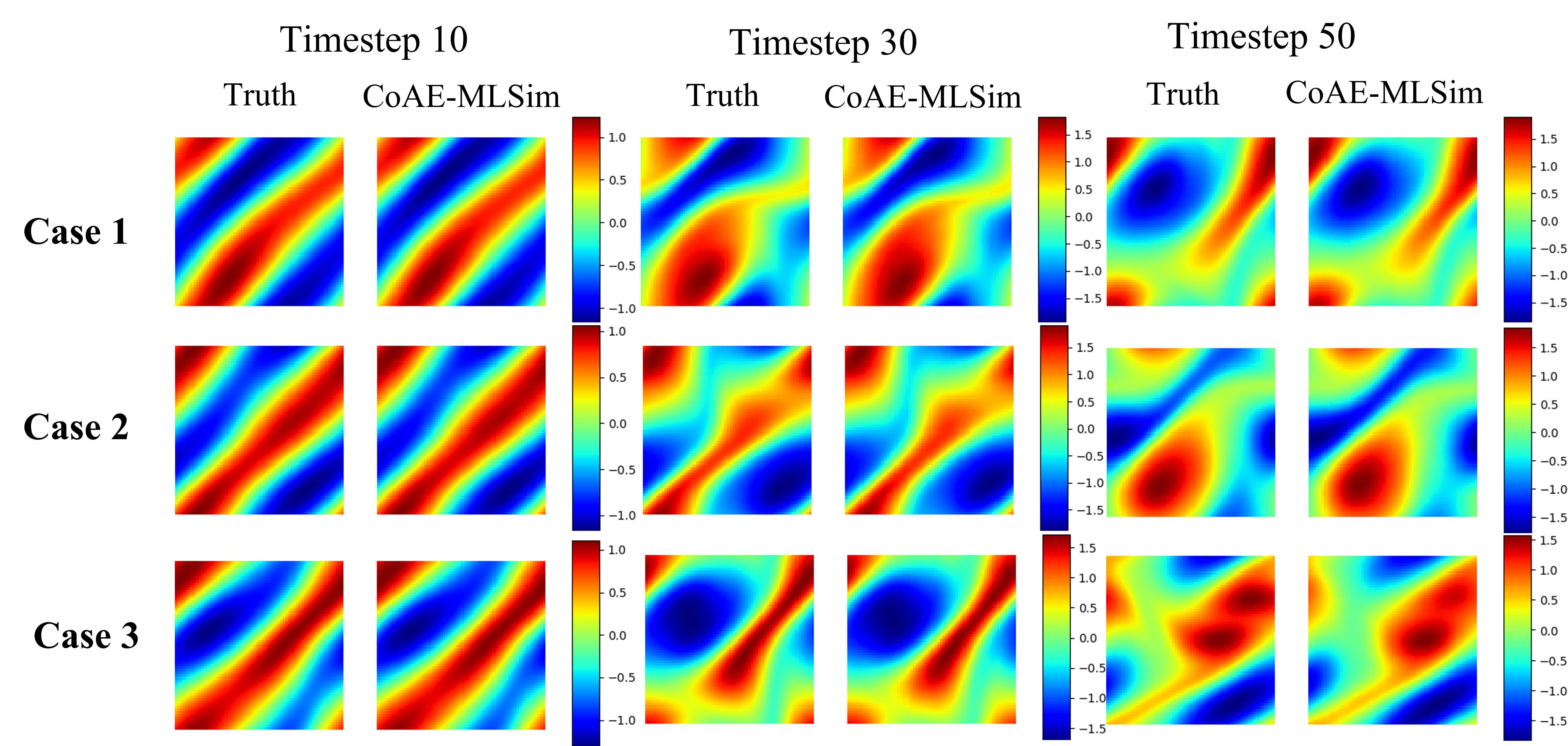}
  \caption{Contour comparisons of vortex decay at different time steps}
  \label{vorticity_decay_contour}
\end{figure}

\begin{figure}[h]
  \centering
  \includegraphics[width=\textwidth]{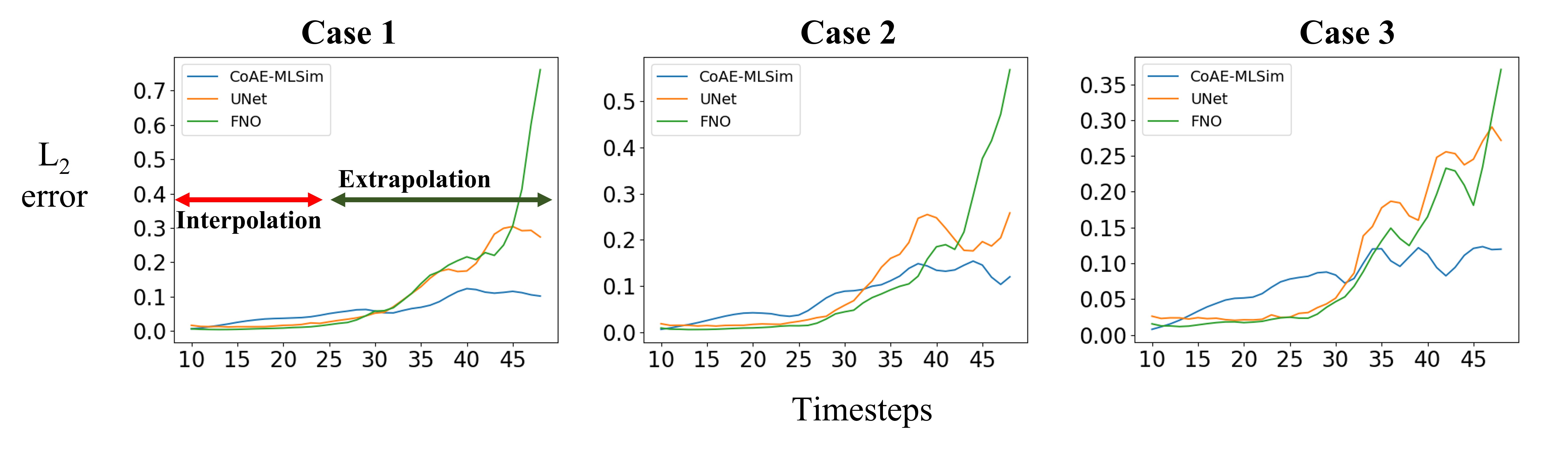}
  \caption{Error accumulation at different time steps}
  \label{vorticity_decay_error}
\end{figure}

\subsubsection{Transient: Flow over a cylinder} \label{appendix:cylinder_transient}

The case setup and geometry is similar to the case presented in Section \ref{appendix:geo} except that the flow Reynolds number is much higher, equal to $200$, in order to induce unsteady phenomenon in the flow, commonly known as vortex street. The governing equations corresponding to this case are presented in \ref{eq8}:

\begin{equation}\label{eq8}
\left. \begin{array}{ll}  
\mbox{\textbf{Continuity Equation:} } \quad\quad\quad\quad\quad\quad\quad \quad\quad\displaystyle\nabla . \textbf{v} = 0\quad\quad \quad\quad\quad\quad\quad\quad\quad\quad\\
\mbox{\textbf{Momentum Equation:} } \quad\quad\quad\quad \quad\displaystyle\frac{\partial v}{\partial t} + (\textbf{v}.\nabla)\textbf{v} + \nabla \textbf{p} - \frac{1}{Re} \nabla^2 \textbf{\textbf{v}} = 0\quad\quad\quad\quad\\[8pt]
\mbox{\textbf{Energy Equation in Fluid:} } \quad\quad\quad\quad\displaystyle\frac{\partial T}{\partial t} + ( \textbf{v}.\nabla)\textbf{T} - \nabla .\left( \alpha\nabla\textbf{T}\right) = 0\quad\quad\quad\quad\quad\\
 \end{array}\right\}
\end{equation}
where, $v={u_x, u_y, u_z}$ is the velocity field in $x, y, z, t$, $p$ is pressure, $T$ is temperature, $Re$ and $\alpha$ are flow and thermal properties.

\begin{figure}[h]
  \centering
  \includegraphics[width=0.8\textwidth]{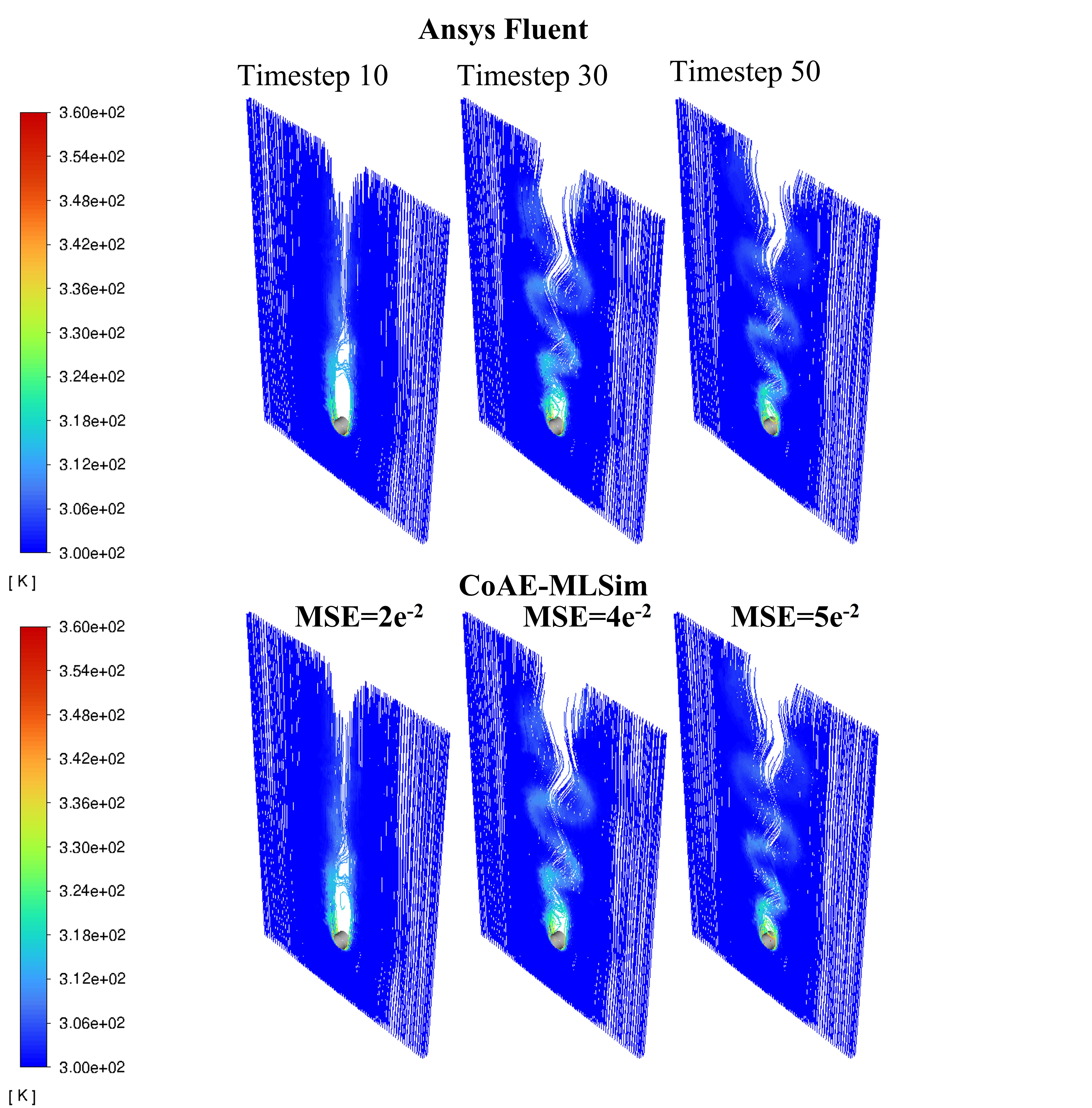}
  \caption{Transient flow comparisons of temperature}
  \label{cyl_trans_temp}
\end{figure}

\begin{figure}[h]
  \centering
  \includegraphics[width=0.5\textwidth]{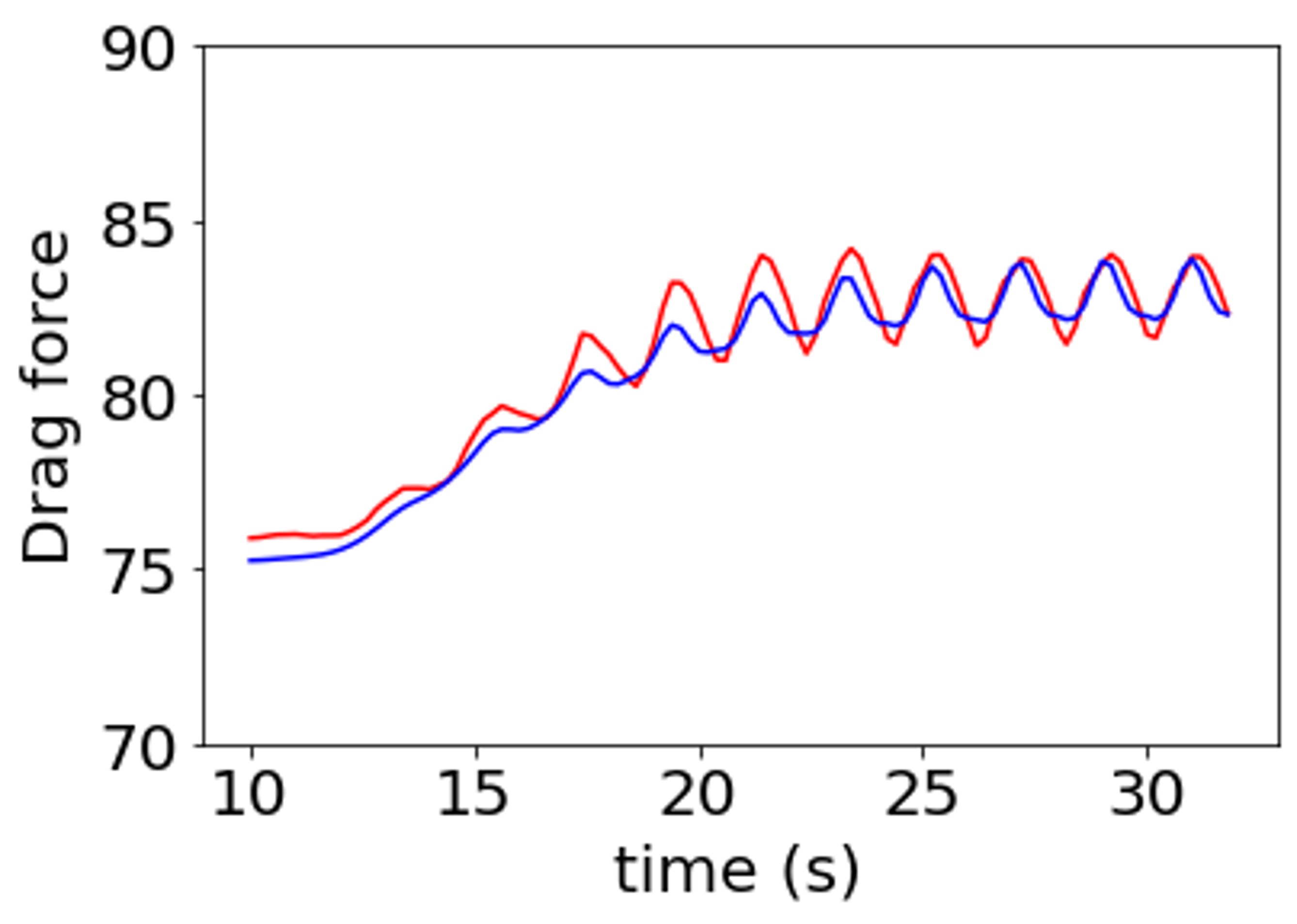}
  \caption{Total drag force comparison}
  \label{cyl_trans_drag}
\end{figure}

Here we provide additional results for temperature profile comparisons in Fig. \ref{cyl_trans_temp}. The comparisons are carried out with Ansys Fluent on cut center planes at $3$ different time steps. The CoAE-MLSim approach captures the transient flow dynamics with an acceptable error in comparison to the Ansys Fluent. Furthermore, in Figure \ref{cyl_trans_drag}, we plot the total surface drag force of the cylinder as a function of time to evaluate the vortex shedding frequency of the flow. The CoAE-MLSim predictions are satisfactory and continues to predict well as the solution progresses ahead in time. 

\subsection{Details for reproducibility} \label{appendix:reproduce}

\begin{figure}[h!]
 \centering
  \includegraphics[width=\textwidth]{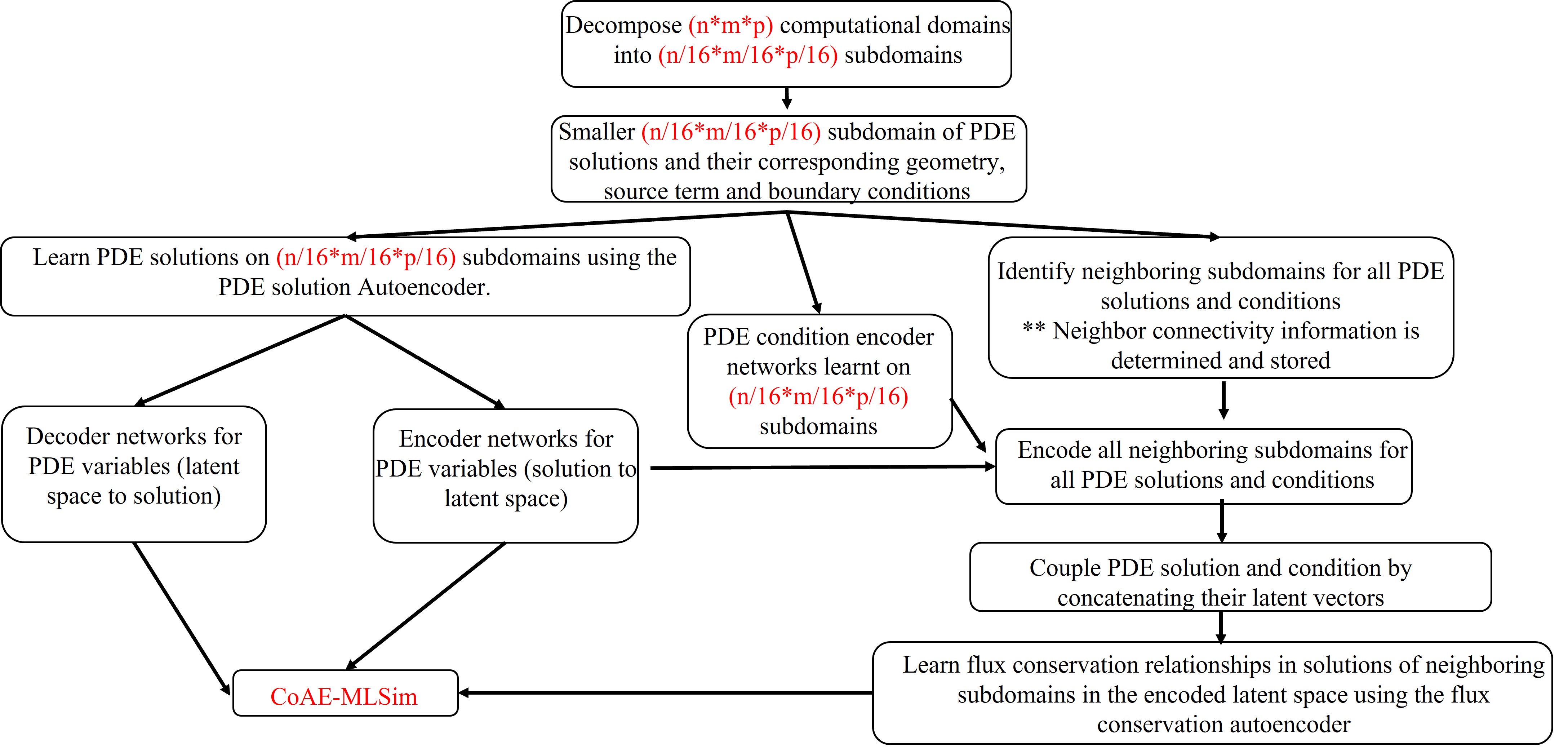}
  \caption{Flow chart for training the CoAE-MLSim approach}
  \label{mlsolver_reproduce}
\end{figure}

In this section, we provide the necessary details for training the CoAE-MLSim. As emphasized previously, the CoAE-MLSim training corresponds to training several autoencoders for PDE solutions, conditions, such as geometry, boundary conditions and source terms and for flux conservation. In Figure \ref{mlsolver_reproduce}, we present a flow chart of the steps that can be followed to train each of these autoencoders. The specific training details and network architectures may be found in Section \ref{appendix:training}. 

For a given set of coupled PDEs, we start with $100$-$1000$ sample solutions on a computational domain with $n, m, p$ computational elements in spatial directions $x, y, z$, respectively. The solutions are divided into smaller subdomains of resolution, $n/16, m/16, p/16$. Each subdomain can has PDEs solutions and PDE conditions associated with it. The PDE solutions are used as training samples to the PDE solution autoencoder to learn a compressed encoding of all the variables on the subdomain. On the other hand, the autoencoders for PDE conditions can be trained with completely random samples, which may or may not be related to sample solutions. Once the PDE solution and condition autoencoders are trained, neighboring subdomains are grouped together and the solution and PDE condition latent vectors on groups of neighboring subdomains are stacked together. These groups of stacked latent vectors are used for training the flux conservation autoencoder. The trained PDE solution, condition and flux conservation autoencoders combine to form the primary components of the CoAE-MLSim. 

The solution algorithm of the CoAE-MLSim has been described in detail in the main body of the paper and may be used for solving for unseen PDE conditions. 

\end{document}